\documentclass{article}
\pdfoutput=1
\usepackage[numbers]{natbib}
\usepackage[preprint]{neurips_2019}

\usepackage{graphicx} 

\usepackage{algorithm}
\usepackage{algpseudocode}

\usepackage{subcaption}

\usepackage{hyperref}

\usepackage[turnoff]{notes}

\usepackage{siunitx}

\usepackage{amssymb}
\usepackage{amsmath}

\usepackage{tikz}
\usepackage{widetext}
\usetikzlibrary{positioning}
\usetikzlibrary{fit}

\title{Learning by stochastic serializations}

\usepackage[bbgreekl]{mathbbol}
\usepackage{amsfonts}
\usepackage{bbm}

\DeclareSymbolFontAlphabet{\mathbb}{AMSb}

\def\Pr{P}

\newcommand{\vect}[1]{\boldsymbol{#1}}

\graphicspath{{./figure/}}

\makeatletter
\renewcommand{\ALG@beginalgorithmic}{\tiny}
\makeatother

\usepackage{xcolor}

\usepackage{xspace}

\newcommand{\strucAlgo}{serialization algorithm\xspace}
\newcommand{\strucLearn}{S-RNN\xspace}
\newcommand{\baseLearn}{RNN\xspace}

\usepackage[spacing]{microtype}

\author{Pablo Strasser\footnotemark \addtocounter{footnote}{-1}\\Department of Business Informatics\\ University of Applied Sciences\\ Western Switzerland \And
Stéphane Armand\\
Willy Taillard Laboratory of Kinesiology\\ Geneva University Hospitals and Geneva University\\
Switzerland \And
Stephane Marchand-Maillet\\
Department of Computer Science\\ University of Geneva\\ Switzerland
\And Alexandros Kalousis\footnotemark\\
Department of Business Informatics\\ University of Applied Sciences\\ Western Switzerland
}

\begin{document}
%
%
%
%
%
\maketitle
\renewcommand{\thefootnote}{\fnsymbol{footnote}}
\footnotetext{Also member of the Department of Computer Science of the University of Geneva}

\begin{abstract}
Complex structures are typical in machine learning.
Tailoring learning algorithms for every structure requires an effort 
that may be saved by defining a generic learning procedure adaptive to any complex 
structure. In this paper, we propose to map any complex structure onto a generic form, 
called serialization, over which we can apply any sequence-based density estimator. We 
then show how to transfer the learned density back onto the space of original structures.
To expose the learning procedure to the structural particularities of the original 
structures, we take care that the serializations reflect accurately the
structures' properties. 
Enumerating all serializations is infeasible.
We propose an effective way to sample representative serializations from  the
complete set of serializations which preserves the statistics of the complete set.
Our method is competitive or better than state of the art learning algorithms that have 
been specifically designed for given structures. In addition, since the serialization 
involves sampling from a combinatorial process it provides considerable protection from 
overfitting, which we clearly demonstrate on a number of experiments. 
\end{abstract}

\vspace{-0.3cm}
\section{Introduction}
\vspace{-0.3cm}

Many learning problems are defined over complex instance structures, e.g. learning instances can be sets, trees, sequences etc. One typical approach to such problems is the so-called propositionalisation, \cite{lavrac2001relational}, in which one maps such complex learning instances to vectorial representations, potentially losing discriminative information along the way. Yet another approach is to develop learning algorithms tailored to the representation particularities of any given problem, preserving in that manner all information, at the cost of significant conceptual and development effort. 

Instead, we propose to decouple learning from the structural specificities of the learning instances.  To do so, we define an informed, randomized, mapping from any given complex-structured instance onto {\em multiple} and {\em equivalent} sequences.
We then learn over the space of sequences and map back the result of the learning onto the space of the original instance structures. When mapping a complex instance structure onto a sequence, we must retain the specificity of the original structure in order to guarantee a revertible mapping
and preserve its properties in order to learn correctly. We do so by carrying over the mapping a set of constraints and properties of the original instance structure 
to the sequences over which we learn.

Our approach opens a sound and systematic way to perform learning over arbitrary complex instance structures,
and allows us to directly use any learning algorithm defined over sequences to do the learning. 
The fact that we map the instances to multiple and equivalent sequences over which we learn brings 
significant advantages when it comes to overfitting avoidance. We experiment with 
generative and discriminative learning settings over complex structures for a variety 
of problems and complex structures.

\vspace{-0.3cm}
\section{Related Work}
\vspace{-0.3cm}

The recent surge on generative modeling has seen the development of generative methods that can learn, implicit or explicit, distributions over complex instances and sample from them. 
We have applications of generative modeling in problems where the learning instances are two dimensional structures, e.g. 
images \citep{PixelRNN,GAN,GeneratifVAE,VAERezende,VAEKingmaW13}, 
sequences for speech \citep{WaveNet}, text \citep{GravesSequence}, translation \citep{Seq2SeqSutskeverVL14},
and graphs, e.g. drug modeling \citep{moleculeDuvenaud}.

Of more interest to our work are generative models for structures such as sets \citep{SetVinyalsBK15,Deep_Sets} and trees \citep{alvarez-melis_tree-structured_2016,TreeZhangLL16,TreeDongL16,TreeLiuXGGLW11,TreeZhouLCXLCH17}. Such models incorporate the specificities of the original instance structure on how they factorize the generative distribution to a product of conditionals. The factorization controls the dependencies between the components of a given learning instance, ensuring that the inherent properties of the original instance structure are preserved. 
Examples of structural properties that can be expressed via the factorization of the conditionals are order independence and invariance to invertible transformations of the conditioned variable \citep{RedOlivaDPXS17,NadeUriaCGML16}.
These properties are used to express invariances of specific complex structures such as invariance to re-ordering of conditioning for sets \citep{SetVinyalsBK15,Deep_Sets}, invariance to re-ordering of siblings for trees, and invariance to relabeling of nodes for graphs.

In our work, we take a constructive modeling approach over the mapped sequences ({\em serializations}), 
we transfer the invariance properties of the original complex structures onto constraints in the procedure for constructing these serializations.
\note[Alex-NIPS19]{Previous sentence is problematic. What do we mean by the "model for constructing these serializations"? Is this 
your actual serialisation procedure? if yest it might be better not to say model. 
Moreover what do you mean by "constructive modelling approach over the mapped sequences"? this latter
seems indeed to be the learning part.}
We model structure invariance via states representing the relevant information that the system has acquired at a given step of the generative process.  
The inherent properties of the original structure are thus expressed by which partial serializations are represented by the same state or not.  For example, the invariance to re-ordering is expressed by the fact that building a given sub-structure by following two different orderings leads to the same state. The specific representation of a state is domain-specific and also allows to incorporate further information about the inherent properties of the original structures.  \note[Removed]{Our approach falls between the full modeling of the complex structures in a domain-specific way and density estimations over serializations/sequences in which one considers no invariance properties.}


Of direct relevance to our work is the Grammar Variational Autoencoder  (GVA) \citep{kusner_grammar_2017} which learns generative models over arbitrary complex structures. 
There structures are described as sequences of production rules of a context-free grammar, sequences over which a standard Variational Autoencoder is then trained. The original instances are reconstructed by predicting the sequence of production rules.
The GVA can be viewed as a special case of our framework since a context-free grammar can be transformed into a state-transition function. In the latter, a state contains a representation of what has been parsed so far, and transitions correspond to the application of the different production rules.
Our framework makes explicit the notion of a {\em state} and uses it to enforce constraints on the probabilities of occurrences of training serializations.

Works in the area of graph neural networks \citep{battaglia_relational_2018,gilmer_neural_2017} study and propose models based on the transfer of information (messages) between nodes and edges. Our method has a similar motivation, it transfers information only to the parts of the model that need it. We have chosen to use a simpler architecture, sequences instead of graphs, at the cost of additional preprocessing; nevertheless our method can be generalized to graphs. The question though of whether the additional complexity would bring performance improvements is open and warrants investigation. 

\note[Removed]{This representation will be used in two ways. First it will be used by ensuring that the empirical distributions of the serializations generated from the original training instances are consistent with the inherent properties of the structure ensuring for example that for an order invariant data structure the probability of the following elements of the sequences does not depend on the order thus ensuring that the data used to train the sequence generator model does not contain wrong inherent properties.
Doing this will indirectly ensure that the hidden states of two different serialization of the same instance are equivalent\footnote{We consider two hidden states equivalent if the output of the network using both values of the hidden states are the same, ie. there is no observable effect when changing the hidden state.} because the target of both networks are the same. This is similar to what is done in orderless NADE\citep{NadeUriaCGML16} to impose order invariant for vector density estimation however done on more complex structures.
Secondly, we propose a regularized version of our framework for RNN like algorithm which in addition to the serialization provide a matrix indicating which hidden state should be the same to respect the inherent properties of the structure.
This constraint was used in TwinNet \citep{twinnet} to enforce the consistency  for the specific data structure of sequence data given in the forward order and backward order. Imposing constraints on the hidden state of an RNN transforms the linear structure of the RNN into a general graph structure.}

\vspace{-0.3cm}
\section{Method}
\vspace{-0.2cm}
\label{Method}


We are given a space $\mathbb{X}$ of complex, structured, learning instances, equipped with an unknown probability distribution $\Pr_{\mathbb X}$, and a training set $\mathcal{X}=\{x_{1},\ldots,x_{n} | x_{i}\in \mathbb{X}\}$ of $n$ instances sampled i.i.d from  $\Pr_{\mathbb X}$. Our goal is to a learn a model  $\Pr_{\mathbb X,\vect \phi }$ of $\Pr_{\mathbb X}$, where $\vect \phi$ is the set of model parameters.
\note[Removed]{In addition, we are given domain specific information about the structure of the probability distribution $\Pr_{\mathbb X}$, which allows us to constrain the learned probability distribution  $\Pr_{\mathbb X, \vect \phi}$ to a family of probability distributions $\mathcal{M}$.}

\subsection{The space of serializations}

To learn the probability distribution over the space of the original complex structures $\mathbb{X}$, we first map structured instances $x\in\mathbb{X}$ onto sequences $a \in \mathbb{A}$. $\mathbb{A}$ is the space of sequences of finite but unknown length over some finite lexicon $\mathbb{B}$; the latter is given by the domain of the problem. We denote by $\mathcal A\subseteq\mathbb{A}$ the set of sequences generated by taking the maps of the training instances $x \in \mathcal X$. We learn a probability distribution, $\Pr_{\mathbb A, \vect \phi}$, over $\mathbb{A}$, by training an RNN with $\mathcal A$. We use the learned distribution to construct $\Pr_{\mathbb X, \vect \phi}$ in the original space.
We call our proposal \strucLearn (as Structural RNN) since it brings additional structural information into a 
RNN-based learning of sequences.


Since $\mathcal{A}$ is the only access the learner has to the structure particularities of the specific problem and their constraints, 
it is critical that we tranfer to it as much information as possible from the original structures.
To carry over the constraints from $\mathbb{X}$ to $\mathbb{A}$ 
we map every $x\in\mathcal{X}$ onto multiple sequences $a_j$, {\em serializations} of $x$, exhibiting the 
invariance properties of the original structure $x$.
We denote by $a_j = [a_j^1, a_j^2, \dots , a_j^T] \in\mathbb{A}$ a serialization of $x$. Its elements are  $a_j^i \in \mathbb B$; 
$T$ is the length of $a_j$. We discuss later the serializations and the {\em \strucAlgo} we use. For now, we simply note that the 
\strucAlgo parses the complex structure and produces in a sequential manner a serialization $a_j$. 
For example, a set $x=\{{\sf A},{\sf B},{\sf C}\}$ can be mapped onto any of its multiple serializations, say $a_1= [{\sf A,B,C}]$, $a_2= [{\sf A,C,B}]$, $a_3= [{\sf B,C,A}]$, etc, thus exhibiting order invariance.
A {\em partial serialization} of $x$ is a subsequence $a_j^{[1:d]}=[a_j^1, a_j^2, \dots , a_j^d ], d \leq T$. 

Invariance properties on $x$ will be mapped onto invariance to local or global reordering in $a_j$ (eg ``swapping elements in the serialization of a set doesn't matter'') and/or conditioning the occurrence of subsequences in $a_j$ to its preceding subsequences (eg ``if 'A' has occurred in the partial serialization, it will not appear anymore since elements occur only once within a set'').

The \strucAlgo defines a mapping from an original instance $x$ onto a stochastic process whose sampled realizations will create the serializations $a_j$, resulting in $\mathcal A$.
This mapping (\strucAlgo) comes from domain knowledge and must be revertible, i.e a serialization $a_j$  generated from instance $x$ reconstructs/de-serializes to $x$ and only $x$.
%
%
%
%
%
%
%

In the next section, we discuss the mapping from $\mathbb{X}$ to $\mathbb{A}$ and how from a probability 
distribution learned over $\mathbb{A}$ we can extract a corresponding distribution on $\mathbb{X}$;
we impose no constraints on the serialisations and the probability distribution on $\mathbb{A}$. In section \ref{ProbStructure} 
we show how to impose structure constraints on the constructed sequences that also constrain the feasible
probability distributions on $\mathbb{A}$ to finally construct $\mathcal{A}$. Thus sections 
\ref{Manifold} and \ref{ProbStructure} describe only how to generate sequences which carry over the appropriate invariances
of the original structures; 
they make no statement about the learning 
model. In section \ref{Regularizer} we define a regulariser that imposes on the learning model the same structural constraints 
as the ones we use to generate the sequences, ie the regulariser brings into the model the same
inductive bias we used to generate the sequences.

\subsection{Serialization with no structural constraints}
 \label{Manifold}
 
 We assume that $\mathbb{A}$ is a probability space equipped with a distribution $P_{\mathbb A}$, of 
which an estimate $P_{\mathbb A, \vect \phi}$ we learn by applying an RNN on the $\mathcal A$ set.
Here, we describe our abstract model for the creation of a relevant training set $\mathcal A$ and
how to generate a distribution $P_{\mathbb X, \vect \phi}$ from $P_{\mathbb A, \vect \phi}$.

 We assume $\mathbb{X}$ to be measurable and define the random variable $X : \mathbb{A} \rightarrow \mathbb{X}$.
 We install a probability distribution $P_{\mathbb X, \vect \phi}$ over the original space $\mathbb{X}$ by pushing 
forward the distribution $P_{\mathbb A, \vect \phi}$, learned in the serialization space $\mathbb{A}$, along $X$.
 The r.v. $X$ classically allows us to compute probabilities over $\mathbb{X}$ by:
 \begin{equation}\label{defProb}
	 \Pr_{\mathbb X, \vect \phi}(x) \triangleq   \Pr_{\mathbb A, \vect \phi}(\{a_j\in\mathbb{A}| X(a_j)=x\}) 
				        =  \Pr_{\mathbb A, \vect \phi}(X^{-1}(x)) = \sum_{a\in X^{-1}(x)}\Pr_{\mathbb A, \vect \phi}(a)
 \end{equation}
 Following our earlier description, $X$ is the de-serialization procedure mapping a serialization $a_j$ to the original data $X(a_j)=x$.  In turn, $X^{-1}$, as a serialization process, represents the stochastic process sampling from the set of all possible serializations of $x$, $X^{-1}(x)=\{a_j\in\mathbb{A}| X(a_j)=x\}$.
 Serializing via a particular \strucAlgo $X^{-1}$, and therefore choosing a subset of all possible serializations, creates a bias in the representation of $x$ within $\mathcal{A}$ that we need to account for to maintain accurate learning.
 To address this bias, we introduce an abstract structure on the space $\mathbb{A}$, which allows us to discriminate between equivalent sequences in $\mathcal{A}$. We call this additional structure {\em properties}.
 
We define a random variable $O : \mathbb{A} \mapsto \mathbb{O}$, where $\mathbb{O}$ is a measurable space of {\em properties}.
Using $O^{-1}$, we map these properties onto $\mathbb{A}$.
An illustrative example of such properties on sequences is ``elements of the sequence are in alphabetical order''. Outcome $o\in\mathbb{O}$ are properties that apply to sequences in $\mathbb{A}$ in general and will help characterize serializations as follows.
A sequence $a_j\in\mathbb{A}$ may be a serialization of $x\in\mathbb X$ (ie $a_j \in X^{-1}(x)$) and/or bear property $o\in\mathbb O$ (ie $a_j \in O^{-1}(o)$).
Thus, given an original instance $x$ and a property $o\in\mathbb{O}$ on sequences, the set of serializations of $x$ 
bearing property $o$ is $O^{-1}(o)\bigcap X^{-1}(x)$.

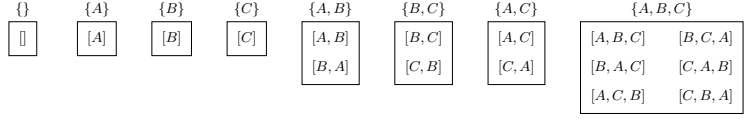
\begin{figure}[hbt]
	\small
	\begin{center}
		\scalebox{0.6}
		{
		\begin{tikzpicture}
		\node(emptySequence){$[]$};
		\node(emptySet)[fit=(emptySequence), draw,label={$\{\}$}] {};
		\node(aSequence)[right=of emptySet]{$[A]$};
		\node(aSet)[fit=(aSequence), draw,label={$\{A\}$}] {};
		\node(bSequence)[right=of aSequence]{$[B]$};
		\node(bSet)[fit=(bSequence), draw,label={$\{B\}$}] {};
		\node(cSequence)[right=of bSequence]{$[C]$};
		\node(cSet)[fit=(cSequence), draw,label={$\{C\}$}] {};
		\node(abSequence)[right=of cSequence]{$[A,B]$};
		\node(baSequence)[below=0.1 of abSequence]{$[B,A]$};
		\node(abSet)[fit=(abSequence)(baSequence), draw,label={$\{A,B\}$}] {};
		\node(bcSequence)[right=of abSequence]{$[B,C]$};
		\node(cbSequence)[below=0.1 of bcSequence]{$[C,B]$};
		\node(bcSet)[fit=(bcSequence)(cbSequence), draw,label={$\{B,C\}$}] {};
		\node(acSequence)[right=of bcSequence]{$[A,C]$};
		\node(caSequence)[below=0.1 of acSequence]{$[C,A]$};
		\node(acSet)[fit=(acSequence)(caSequence), draw,label={$\{A,C\}$}] {};
		\node(abcSequence)[right=of acSequence]{$[A,B,C]$};
		\node(bacSequence)[below=0.1 of abcSequence]{$[B,A,C]$};
		\node(acbSequence)[below=0.1 of bacSequence]{$[A,C,B]$};
		\node(bcaSequence)[right=0.5 of abcSequence]{$[B,C,A]$};
		\node(cabSequence)[below=0.1 of bcaSequence]{$[C,A,B]$};
		\node(cbaSequence)[below=0.1 of cabSequence]{$[C,B,A]$};
		\node(abcSet)[fit=(abcSequence)(bacSequence)(acbSequence)(bcaSequence)(cabSequence)(cbaSequence), draw,label={$\{A,B,C\}$}] {};
		\end{tikzpicture}
		}
		\caption{Serialising set structures. Rectangles below each set indicate the equivalent serialisations.}
		\label{SetShema}
	\end{center}
\end{figure}

Let us demonstrate these notions using sets as our structure example. In figure \ref{SetShema} we show all possible serializations of sets with up to three elements. In the rectangle below each set we give all its serialisations. The properties correspond to an ordering which allows us to distinguish equivalent but different serializations of a given set (structure). In the example, all six 
serializations of $\{A,B,C\}$ are equivalent and are differentiated only by their ordering.

We want to learn the density of a given structure irrespective of the serializations choices (properties). 
We could marginalize out the properties. However this would be intractable for structures 
with large equivalence classes.  Instead, we adapt equation \ref{defProb} and integrate the choice of the 
serialization (its probability), controlled by the $o$ variable: 
 \begin{equation}\label{FracProb}
	 \Pr_{\mathbf X, \vect \phi}(x)=\frac{\Pr(o,x)}{\Pr(o|x)}=\frac{\sum_{a\in X^{-1}(x)\cap O^{-1}(o)}\Pr_{\mathbb A, \vect \phi}(a)}{\Pr(o|x)}\quad \forall o\in \mathbb{O}
 \end{equation}
 $\mathbb{O}$ structures  $\mathbb{A}$ and helps us count the distinct (but equivalent) serializations of a given $x$;
it comes from domain knowledge. For example for a set with $n$ elements we know that we have $n!$ orderings of its elements, 
thus $n!$ distinct but equivalent serialisations.

 
 
We denote by $\mathbb{F}\subseteq 2^{\mathbb{X}}$ the set of events of $\mathbb{A}$.  We see the stochastic process installed by $X^{-1}$  as a sampling strategy defined by an arbitrary measure $\mu:\mathbb{F}\mapsto \mathbb{R}_{>0}$.
This arbitrary measure $\mu$ indicates the importance we put on a serialization with respect to other possible serializations.
\note[Alex-NIPS19]{You use two terms serialisation scheme vs other possible serialisations, are you refering to the same type of object? not clear.}
Hence abstractly, a run of the \strucAlgo computes a set of serializations $X^{-1}(x)$ and picks one according to the weight measure $\mu$ for feeding the training set $\mathcal{A}$.
$\mu$ therefore comes as a support to the calculation of our normalizer $P(o|x)$ as
 \begin{equation}\label{normalizer}
	 \Pr(o|x)\triangleq\frac{\mu( X^{-1}(x)\cap O^{-1}(o))}{\mu(X^{-1}(x))}
 \end{equation}
In the set example above if we treat all orderings as equiprobable we have 
$\mu(X^{-1}(\{A,B,C\}))=3!$ and $\mu( X^{-1}(\{A,B,C\})\cap O^{-1}([B,A,C]))=1$.
In the next section we will exploit and modify this abstract modeling to enforce constraints on 
the serializations that reflect the invariance properties of the original structure. In practice, 
unless we have a good reason to do otherwise we will choose the uniform distribution for $\mu$.

To compute $\Pr_{\mathbb X, \vect \phi} (x)$, we need $\Pr_{\mathbb A, \vect \phi}(a)$ (eq.~\ref{FracProb}). \note[Removed]{$\Pr(o|x)$ has an exact solution which we get from \color{red}????\color{black}.} \note[Alexandros]{Where from does the exact solution for  $\Pr(O=o|X=x)$ come from?} We fit $\Pr_{\mathbb A, \vect \phi}$ by maximum likelihood on $\mathcal A$.
 

\subsection{Serializations with structural constraints}
 \label{ProbStructure}

 We place serializations into $\mathcal A$ on the basis of their global occurrences 
(ie as complete serializations) and the sampling distribution $\mu$. However, we do 
possess additional information on the structure which we will incorporate in our 
sampling procedure.
 
In the set example, since a set is order invariant, serializations $a_3=[{\sf B,C,A}]$ and $a_4=[{\sf C,B,A}]$ are equivalent, 
and so are their partial serializations $a_3^{[1:2]}=[{\sf B,C}]$ and $a_4^{[1:2]}=[{\sf C,B}]$.
 We can explicitly model this equivalence by constraining the conditional probability of the next element of two serializations, given equivalent partial serializations, to be the same, i.e. $P_{\mathbb A, \phi}(a_3^3=b|a_3^{[1:2]})=P_{\mathbb A, \phi}(a_4^3=b|a_4^{[1:2]})\quad \forall b\in \mathbb B=\{\sf A,B,C\}$. More generally, given two serializations $a_i$ and $a_j$, the $t$-length partial serializations of which are equivalent, we require that: 
\begin{align}\label{equivalence}
	\Pr(a_i^{t+1}=b|a_i^{[1:t]}) & = \Pr(a_j^{t+1}=b|a_j^{[1:t]})\quad \forall b\in \mathbb B
\end{align}

Equation \ref{equivalence} transfers the structural invariances of the original instances $x \in \mathbb X$ to 
the serialisations $a \in \mathbb A$ produced from them. 
The constraints are enforced by identifying equivalent partial serialisations of $a_{i}$ and $a_{j}$ and ensuring that the probability 
distribution of the next element is the same.
To do so we define a state space $\mathbb{S}$ and map sequences $\mathbb{A}$ on to it 
so that equivalent partial sequences have the same state.
$\mathbb{S}$ is equipped with a transition function $f: \mathbb{S} \times \mathbb{B} \rightarrow \mathbb{S}$ governing the construction of a sequence via its equivalent states.
Hence, a serialization $[a^{1},\ldots,a^{t},\ldots, a^{T}]$ is represented by the sequence of states $s^0,\ldots,s^t,\ldots, s^T$ produced by the recurrence:
\begin{equation}\label{Transition}
s^{t+1}=f(s^t,a^{t+1})\quad \forall 0\leq t< T
\end{equation}
where $s^0$ is an initial state representing an empty sequence in $\mathbb{A}$, and therefore an empty object in $\mathbb{X}$ (eg a graph with no node or an empty set).
At any step, a partial sequence $a^{[1:t]}=[a^{1},\ldots,a^{t}]$ is represented by state $s^t$.
Hence,
\begin{equation}
\label{Prob:state}
\Pr(a^{[t+1:T]}|a^{[1:t]})=\Pr(a^{[{t+1}:T]}|s^t)
\end{equation}
By combining equations \ref{Transition} and \ref{Prob:state}, and imposing at step $t$ that the state $s_i^t$ of partial sequence $a_i$ is the same as the state $s_j^{t}$ of partial sequence $a_{j}$, if the two partial sequences are equivalent, we model the equivalence relationship in equation \ref{equivalence} in the state space:
\begin{align}\label{equivalencestate}
	\Pr(a_i^{t+1}=b|s_i^t)  = \Pr(a_j^{t+1}=b|s_j^t)\quad \forall b\in \mathbb B\quad s_j^{t}=s_i^{t}
\end{align}

Modeling with states enables the incorporation of structural constraints on the serializations.
Such constraints create correlations between (sub-)serializations and prevent us from sampling 
serializations at once, as we did in the previous section.
Even different instances $x$ may share substructures and thus share equivalent partial serializations.
We thus need to enforce equivalence constraints across different instances.
To do so we sample at the level of the serialization element.
We adapt our sampling measure $\mu$ to reflect these equivalence constraints.
In other words, we adapt $\mu$ to express how the next element $a^{t+1}$ is sampled from $\mathbb B$ with respect to the state $s^t\in\mathbb S$.
We redefine the measure as $\mu: \mathbb{S}\times \mathbb{B} \rightarrow \mathbb{R}_{>0}$ to provide a measure over the joint set of states ($s^t$) and the lexicon from where $a^{t+1}$ will be sampled. The $\mu$ measure allows to prioritize orderings.
Because we define $\mu$ on the states and not on the past sequences, we automatically ensure that our sampling follows the constraints.

In practice, defining an appropriate sampling strategy is difficult but this leads us to an interesting algorithmic solution.
We give in Algorithm \ref{alg:Sampling} of the appendix a procedure to efficiently sample a serialization, focusing on  the structural constraints and compensating for the bias coming from the specificity of the base serialization algorithm $X^{-1}$.
In a nutshell, given a base set of serializations $\mathcal{A}_{\rm all}=X^{-1}(x)$, the procedure samples, element by element, a serialization guaranteed to come from the base set according to the statistics of the base set and $\mu$.
At each time step $t$, the procedure stores in the set $\mathcal L$ all elements $a_j^{t}$ as possible next elements for the currently reconstructed sequence $a_{\rm sample}^{[1:t-1]}$.
The next element $a^{\rm next}$ is then sampled from $\mathcal L$ using $\mu$ and $s$ and concatenated to $a_{\rm sample}^{[1:t-1]}$ (ie $a_{\rm sample}^t=a^{\rm next}$).
In order to preserve the consistency of the sampled serialization, all serializations in $\mathcal{A}_{\rm all}$ not having element $a^{\rm next}$ as $t$th element are removed from $\mathcal{A}_{\rm all}$. This is repeated until the special $eos$ token is sampled.
The final training set $\mathcal{A}$ is obtained by sampling the serialization $\mathcal{A}_{\rm all}$ for every training instances $x$.


Note that for example, in the simple case of sets of size $n$ with uniform sampling, this decimation (equivalent to saying that every element may appear once only and at any position) will lead to every serialization (with no restriction on their property) having a probability of $\frac{1}{n!}$, which is consistent with the reality.
On the other hand, choosing a sampling strategy that always samples in alphabetical order will lead to a single serialization (with no restriction on their property) having a probability of $1$ which is also consistent with reality.
So, except for biasing the learner for a particular serialization order, there is no reason to use a different distribution than uniform for $\mu$. Note however that the final sampling strategy described in algorithm \ref{alg:Sampling} is not uniform. Only the distribution of the next element given the current state is.
States are then updated following equation \ref{Transition}.
This procedure therefore makes very efficient the creation of an unbiased training set $\mathcal A$ informing the learner about the structural constraints within $\mathbb X$.




%
%
%

\subsection{Regularized learning}
\label{Regularizer}
Having constructed $\mathcal A$ 
we use a RNN to learn $P_{\mathbb A,\vect \phi}$.
We can support the RNN learning by defining a regularizer based on the constraints of 
our domain. To this end, we will use the structural constraints we gave in equation \ref{equivalencestate} 
(to guide the generation of serialisations) to define a corresponding regulariser on the states that the RNN
learns over these serialisations. 
We create a binary sparse matrix $\left(C_{jk}^t\right)$ (${j,k=1\dots|\mathcal A|}, {t=1\dots T}$) 
storing state equivalences with the serializations $a_j\in \mathcal A$ (ie $C_{jk}^{t}=1$ iff $s_{j}^t=s_{k}^{t}$).
Let $h_t\in\mathbb{R}^{H}$ be the $H$-dimensional hidden state of the RNN.
The probabilistic model of an RNN is given by:
\begin{equation}
h^{0}=0~~~;~~~
h^{t}=\sigma(W_{hh}h^{t-1}+W_{hi}a^{t})~~~;~~~\Pr(a^1,\ldots,a^T)=\prod_{t=1}^{T} \Pr_{\theta}(a^{t}|h^{t-1})
\end{equation}
where $W_{hh}$ is the hidden-to-hidden weight matrix, $W_{hi}$ the input-to-hidden weight matrix and $P_{\theta}$ is a distribution with parameters $\theta$. 
We will use \strucLearn to learn generative (conditional and uncoditional) 
as well as discriminative models. In the former case the goal is to learn to generate
the complex strucures. The learning objective is:
  \begin{eqnarray}
    L=&-\sum_{j=1}^{|\mathcal A|}\sum_{t=1}^{T}\ln(\Pr_{\theta}(a_{j}^{t+1}|h_{j}^{t}))
    +\lambda \sum_{k,l,t}C_{kl}^{t}\left|\left| \frac{h_{i}^{t}}{||h_{i}^{t}||}-\frac{h_{j}^{t}}{||h_{j}^{t}||}\right|\right|
 \end{eqnarray}
where the first term is the loss and the second is the regulariser. When we learn a conditional generative model 
the loss term will also include a $t=0$ step which will be conditioning the $h_j^1$ hidden state on the conditioning 
variables.  In the discriminative modelling, where the goal is to predict a target value ($y_j$) 
from a complex structure, we use a discriminative loss defined over the last hidden state and the target variable 
replacing the log-likelihood term above with $\ln(\Pr_{\theta}(y_{j}|h_{j}^{T}))$. 

The regulariser enforces exactly the same bias in learning that we used to generate the serialisations. Given the, often, 
combinatorial nature of the sequence generation we will have very large sample sets over which we will train; in such cases 
the utility of regularisation is limited, if any. A fact that we also confirm in our experiments.

\subsection{Recovering the density on the original structures}
\label{Recovery}
Given a trained model we want to compute the probability of an instance $x_{\rm test}$ and obtain $P_{\mathbb X,\vect \phi}(x_{\rm test})$ using equations \ref{FracProb} and \ref{normalizer}.
Additionally, in practice we know what serialization algorithm we use, and also know its properties (ie the properties of the serializations it produces).
We use this knowledge to compute $P(o|x_{\rm test})$ without generating all serializations in $X^{-1}(x_{\rm test})$ and also use the same algorithmic enumeration of serializations as proposed in Algorithm \ref{alg:Sampling} to estimate their specific probability of occurrence.


Hence, given one serialization $a_j$ of $x_{\rm test}$ with property $o_j$, we access its learned probability $P_{\mathbb A,\vect \phi}(a_j)$.
We can extract the number of serializations $a$ bearing the same property $o_j$ and therefore easily compute $P(o_j,x_{\rm test})$.
Similarly, by normalizing serializations for their probability of occurrence, we get $P(o_j|x_{\rm test})$ and therefore are able to compute an estimate of $P_{\mathbb X,\vect \phi}(x_{\rm test})$ via $a_j$.
Generating $m$ equivalent serializations of $x_{\rm test}$ by using $m$ times algorithm \ref{alg:Sampling} to obtain $(a_j)$ with $j=1,\ldots,m$ (each bearing property $o_j$), we can improve the accuracy of the estimate by taking the expectation so we finally get:
\begin{equation}
\Pr(X=x_{\rm test})\approx \frac{1}{m}\sum_{j=1}^{m}\frac{\sum_{a\in X^{-1}(x_{\rm test})\cap O^{-1}(o_{j})}\Pr_{\mathbb A,\vect \phi}(a)}{\Pr(O=o_{j}|X=x_{\rm test})}
\end{equation}

\section{Experiments}
We experiment on a set of learning problems where learning instances have diverse
structures, sets, trees, graphs, multi-variate times-series. We learn generative (conditional and unconditional) models
and discriminative (classification and regression) models and show that \strucLearn achieves comparable performance 
with state of the art baselines that have been specifically tailored to these structures. 

\paragraph{Set problems}
\label{sec:pointcloud-short}
We use \strucLearn in a discriminative manner to solve a classification problem over sets.
We serialize a set to a sequence using a random ordering of the elements of the set.
The task is to classify a 3D model of an object represented 
as an infinite set of 3D points (the surface of the 3D model). 
We use the ModelNet10 and ModelNet40 datasets, \citep{song_3d_2015}, which have different 
objects from ten and 40 different 
object types. After preprocessing of the points, we obtain a set of 6 dimensional feature 
vectors ($\mathbb{B}=\mathbb{R}^6$) from which we randomly 
sample $T$ points. We add a classifier to the last state for the final classification. 
We experiment with and without our regulariser ($\lambda=1, \lambda=0$ respectively) 
and the results are identical, showing, as expected in our setting, that the regulariser cannot bring
performance improvements.
\note[Alex-NIPS19]{Is this indeed a valid conclusion? I would be tempted to say that 
the regularise does nothing. To reach a conclusion like the one above shouldn't you 
really explore the values that the regularisation term takes over the learning and
show that it does or does not do something?}

We summarize the results in table \ref{tab:PointCloudResult}.  With $T=500$ points our 
model achieves the same performance as Deep-Set, specifically designed for set problems. 
\note[Alex-NIPS19]{Add citations to Deep-Set, or is is DeepSets?}
Moving to larger sets ($T=5000$), and 
thus larger sequences, does not improve the performance a rather known fact with RNNs. 
\note[Alex-NIPS19]{Citation to support this known fact.}
For completeness we provide the best results on these datasets obtained by 
RotationNet \citep{kanezaki_rotationnet:_2018}. 
However we should note that the RotationNet's model is not based on the concept 
of a set, it rather uses CNNs on multiple learned views of the objects. 
For a more detailed discussion on the experiments see section 
\ref{Long-Set} in the appendix.


\note[Removed]{
Note that for these datasets there is an infinite amount of possible serializations 
for any given object. This  means that every serialization is almost unique. Interestingly, 
in training, our model classifies almost perfeclty new samples of an object it has already 
seen. However it generalises less well to new objects of the same object type in test time. 
}

\begin{table}
\begin{center}
\tiny
\begin{tabular}{l|c|c}
	\hline
	Algorithm &ModelNet40 &ModelNet10 \\ \hline
	S-RNN $T=500$ $\lambda=1$ &$82\%$&$87\%$\\
	S-RNN $T=5000$ $\lambda=0$ &$82\%$&$87\%$\\
	S-RNN Curiculum&$81\%$&$87\%$\\
	Deep-Set $T=500$&$82\%$&-\\
	Deep-Set $T=5000$&$90\%$&-\\
	RotationNet&$97\%$&$98\%$\\
	\hline
\end{tabular}
\caption{Test accuracy on the set problems. } 
\label{tab:PointCloudResult}
\end{center}
\end{table}


\paragraph{Tree problems}
\label{sec:tree-short}

We experiment with two learning tasks where instances are trees,
ordered (the childrens' order matters), or unordered (order does not matter). 
We serialize the former by traversing their nodes from the root to the leaves;
since here the order is important, we cannot use 
randomization. We serialize unordered trees in the same way but now in addition 
we randomize the childrens' order. We experiment with two learning problems.
In the first we use \strucLearn to learn a conditional generative model that 
generates an unordered tree given its textual description. 
We compare against the DRNN baseline \cite{alvarez-melis_tree-structured_2016}.
We evaluate the conditional tree generation as a node and edge retrieval task using 
precision, recall and F1 score.  For lack of space we give  a complete 
description of the setting and the results in appendix section~\ref{sec:tree}, 
table~\ref{TreeSynth}. \strucLearn outperforms the baseline 
by an important margin for all measures except recall.  
In the second learning problem we use \strucLearn to learn a discriminative model that predicts 
a scalar given a tree.  Here we compare against the results of two Tree Echo State Network variants,  
TreeESN-R and TreeESN-M \citep{DBLP:journals/ijon/GallicchioM13}.  We consider both ordered
and unordered trees. \strucLearn, trained on ordered and unordered trees, gives
better results than TreeESN-R, but performs worse than TreeESN-M. For detailed results and discussion 
appendix section~\ref{sec:tree}. 



\paragraph{Graph/Molecule problems}
\label{sec:molecule-short}
We evaluate \strucLearn on regression problems in which instances are graphs that describe molecules.
We serialise the molecular graph into SMILES strings~\citep{weininger_smiles_2002}.
We experiment with canonical (non-randomized) and non-canonical (randomized) SMILES. The latter is the default serialization we
use in \strucLearn. We experiment on the QM9 dataset from the deepchem benchmark~\citep{Ramsundar-et-al-2019} 
and compare against a number of relevant baselines. In the appendix we also include additional results 
on a set problems defined on the Guacamole dataset~\citep{brown_guacamol:_2018}.
For both datasets we use \strucLearn to learn a discriminative model for regression. 
We did not use the regularizer ($\lambda=0$).
We evaluate the predictive performance reporting the squared Pearson coefficient on the test set. 
Our results, table  \ref{result:QM9}, show that the use of non-canonical 
SMILES outperforms the canonical representation and solves nearly perfectly 
all tasks, with the exceptions of mu. It also outperforms all baselines with the exception of the one that uses the 3D position 
information of the atoms. Note that we do not use such information in the two representations
with which we experimented. For detailed descriptions of the experiments and results see appendix, 
section~\ref{sec:graph}.

\begin{table}
	\tiny
	\centering
	\begin{tabular}{l|c|c|c|c|c|c|c|c|c|c|c|c}
		\hline
		Algorithm                 &mu&alpha&HOMO&LUMO&gap&R2&ZPVE&Cv&U0&u298&h298&g298\\
		\hline
		\strucLearn Non-Canonical SMILES               &0.75&0.993&0.93&0.985&0.97&0.970&1.000&0.995&1.000&1.000&1.000&1.000\\
		\strucLearn Canonical SMILES &0.64&0.994&0.90&0.980&0.95&0.968&1.000&0.994&1.000&1.000&1.000&1.000\\
		Best Baseline w/o 3D coord& 0.76&0.963&0.91&0.977&0.96&0.977&0.988&0.978&0.994&0.994&0.994&0.994\\
		Best Baseline w 3D coord  & 0.97&0.993&0.96&0.988&0.98&0.998&0.999&0.997&0.998&0.998&0.998&0.998\\
		\hline
	\end{tabular}
	\caption{Predictive performance on QM9 dataset, Pearson correlation coefficient with the target. 
}
	\label{result:QM9}
\end{table}

\note[Removed]{
\begin{table}
	\small
	\centering
	\begin{tabular}{l|c|c|c|c|c}
		\hline
		Algorithm                 &logP&mol\_weight&num\_atoms&num\_H\_donors&tpsa\\
		\hline
		\strucLearn Non-Canonical SMILES        &0.999&0.999  &0.999  &0.999&0.999\\
		\strucLearn Canonical SMILES            &1.000&0.999  &0.999  &0.999&0.999\\
		\hline
	\end{tabular}
\caption{Predictive performance on the Guacamole dataset, Pearson correlation coefficient with the target.}
\label{result:Guacamole}
\end{table}
}

\paragraph{Multivariate dynamical systems datasets} \label{sec:dynamical-system-short} 

We explore the performance of \strucLearn on multivariate dynamical systems problem where the instances
are essentially multi-variate time series. We serialise these multi-variate time series variable by variable 
using as features the combination of variable id and value. The order of variable serialisation 
is random. We indicate time advancement with a dedicated symbol. 
We report here the results on a real world dataset, gait, which contains recordings of gait trajectories of people 
with pathological gait. The goal is to learn to generate the evolution of different joint angles in time; we experiment
with different number of angles (1, 2, 4, 8).
In addition to that dataset we also experiment with an artificial dataset generated using 
a known dynamical system; for lack of space we report these results in the appendix.
On the gait dataset we use \strucLearn to learn a conditional and an unconditional generative model.
In the former we generate a multivariate time-series (gait) given some patient specific 
input feature in the latter we generate plausible gaits. 
As baseline we use a standard RNN that has the same architecture as our model. In the unconditional 
generation the \strucLearn outperfrorms the RNN in a significant manner for all the number 
of angles with which we experimented. In the case of conditional generation each model has 
one significant win (table \ref{tab:gaitDataUnSupervisedExpLikelihood-short-tab:GaitSupervisedLikelihood-short}).
We also did experiments in order to study the effect of the regulariser 
(cf. table \ref{tab:simulatedDataSupervisedExpLikelihood} of the appendix). As we also 
saw in the set experiments, its use does not bring significant performance improvement. 
We believe that this happens because the randomization provides a considerable part of 
the information that is also exploited in the regularizer.  For more details on this 
set of experiments see section \ref{sec:dynamical-system} of the appendix.

\note[Shortened-NIPS]{
\begin{table}[htbp]
	\begin{subtable}{0.5\linewidth}
	{
		\small
		\begin{center}
			\begin{tabular}{|c|c|c|}
				\hline
				\# of Angles (k)&\strucLearn&\baseLearn\\ \hline
				&mean&mean\\
				\hline
				2&$\mathbf{-6.8}$&7.3\\
				4&$\mathbf{-66}$&-14\\
				8&$\mathbf{-32}$&83.3\\	
				\hline
			\end{tabular}
		\end{center}
	}
	\caption{}
	\label{tab:gaitDataUnSupervisedExpLikelihood-short}
\end{subtable}

\begin{subtable}{0.5\linewidth}
{
\begin{center}
	\tiny
\begin{tabular}{|c|c|c|c|c|c|c|c|} \hline

Angles&\strucLearn                                                                                     &\multicolumn{2}{|c|}{RNN}\\ \hline
      &mean                                                                                            &mean           &  p-value\\ \hline
1     &-47                                                                                             &$\mathbf{-53}$ & 1\\
2     &$\mathbf{-125}$                                                                                 &-97            &$\mathbf{0}$\\	\hline
\end{tabular}
\end{center}
}
	\caption{}
\label{tab:GaitSupervisedLikelihood-short}
\end{subtable}
	\caption{Test set negative log likelihood on the unconditional (left) and conditional (right) gait generation. 
	Bold indicates performances that are significantly better than the non-bold. }
\vspace{-1cm}
\end{table}
}

\begin{table}[hptb]
	\begin{center}
\tiny
	\begin{tabular}{|c|c|c|}
				\hline
				\# of Angles (k)&\strucLearn&\baseLearn\\
				\hline
				2&$\mathbf{-6.8}$&7.3\\
				4&$\mathbf{-66}$&-14\\
				8&$\mathbf{-32}$&83.3\\	
				\hline
			\end{tabular}
\quad
\begin{tabular}{|c|c|c|c|c|c|c|c|} \hline

\# Angles (k)&\strucLearn      &                    {RNN}\\ \hline 
1     &-47                     &$\mathbf{-53}$    \\
2     &$\mathbf{-125}$         &-97                         \\	\hline
\end{tabular}
	\end{center}
\caption{Test set negative log likelihood on the unconditional (left) and conditional (right) gait generation. 
Bold indicates performances that are significantly better than the non-bold.}
\label{tab:gaitDataUnSupervisedExpLikelihood-short-tab:GaitSupervisedLikelihood-short}
\end{table}

\begin{figure}[hptb]
	\begin{minipage}{0.24\linewidth}
		\centering
		\includegraphics[width=\linewidth]{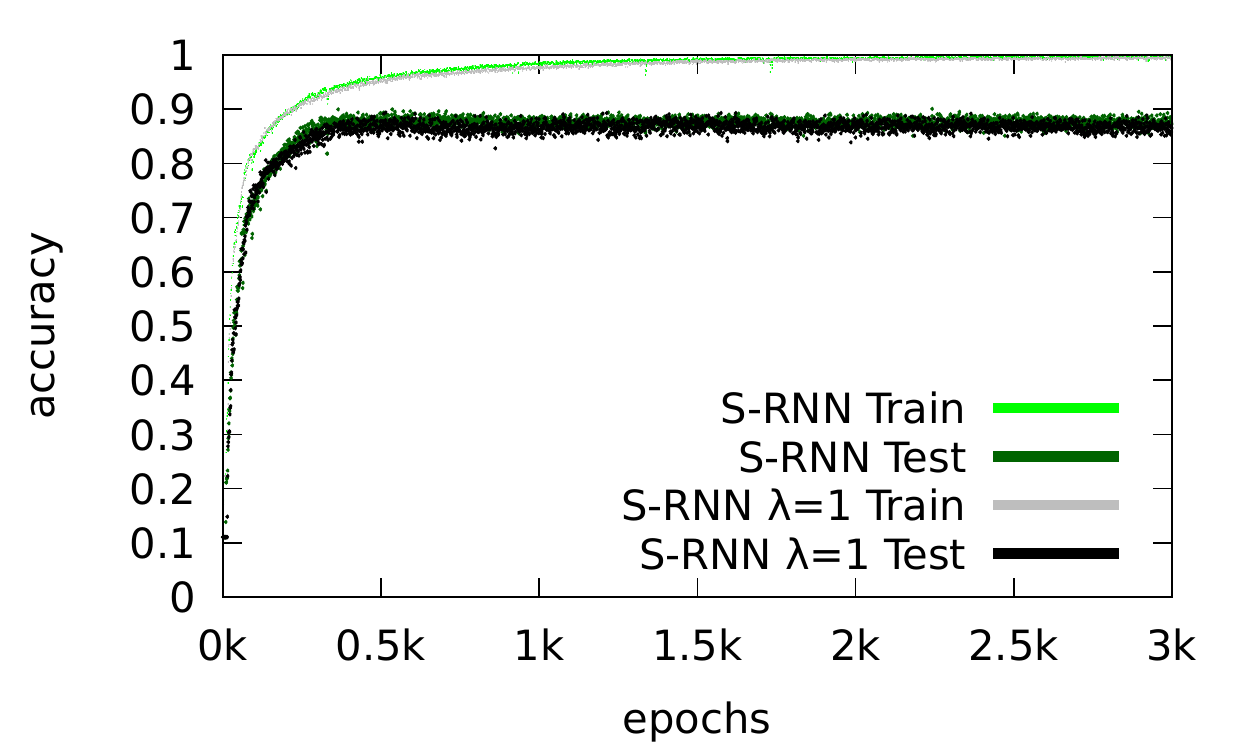} \\
		\tiny Sets
	\end{minipage}
	\hfill
	\begin{minipage}{0.24\linewidth}
		\centering
		\includegraphics[width=\linewidth]{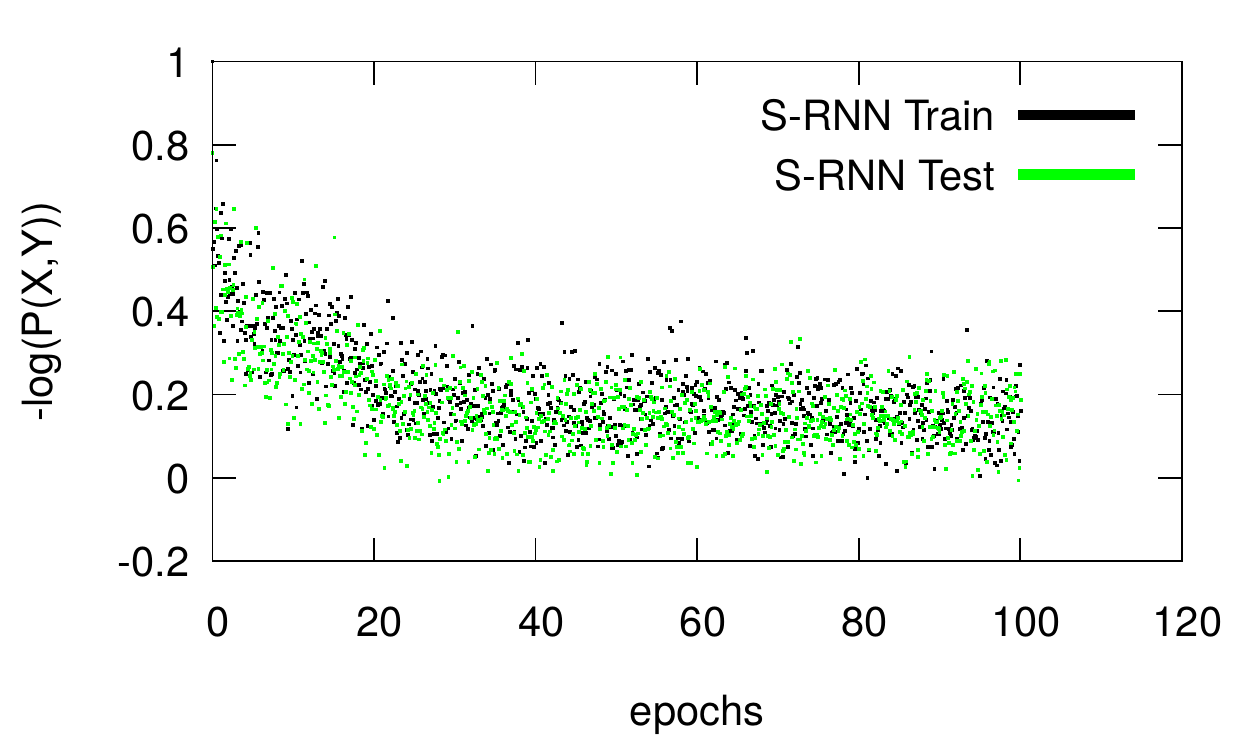} \\
		\tiny Trees 
	\end{minipage}
	\hfill
	\begin{minipage}{0.24\linewidth}
		\centering
		\includegraphics[width=\textwidth]{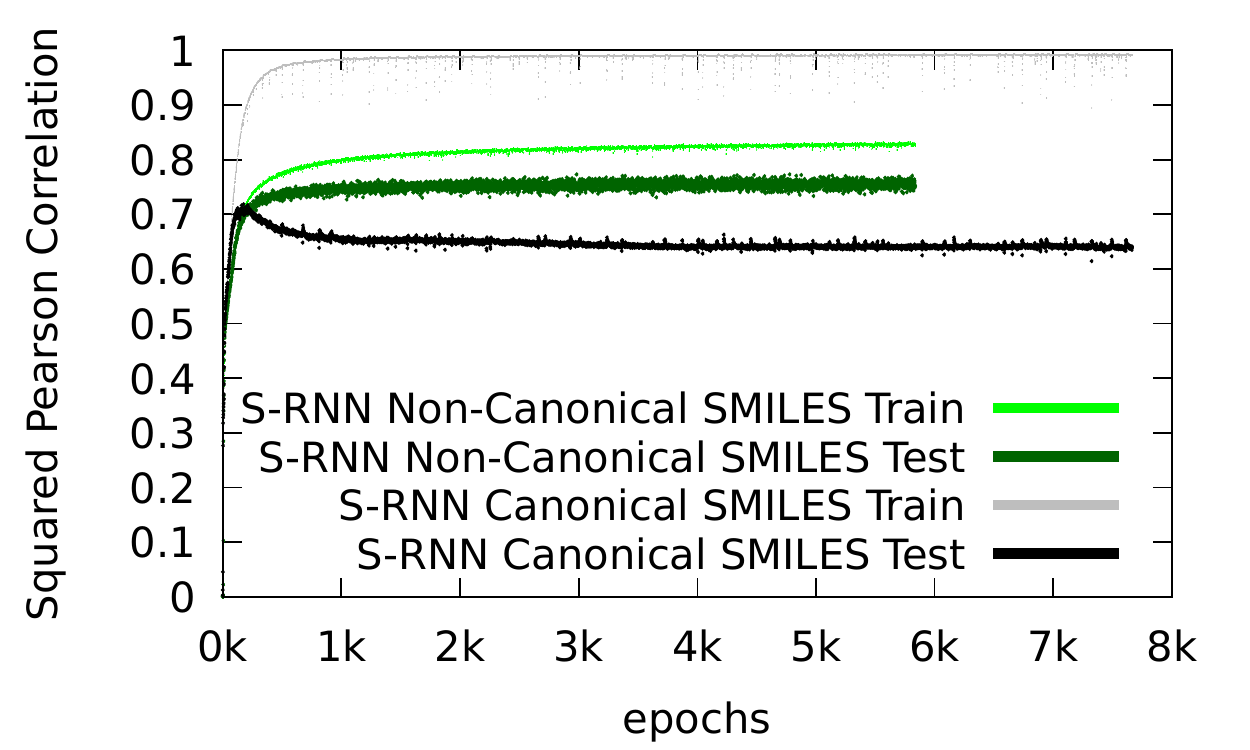} \\
		\tiny Graphs
	\end{minipage}
	\hfill
    \begin{minipage}{0.24\linewidth}
    	\centering
    	\includegraphics[width=\linewidth]{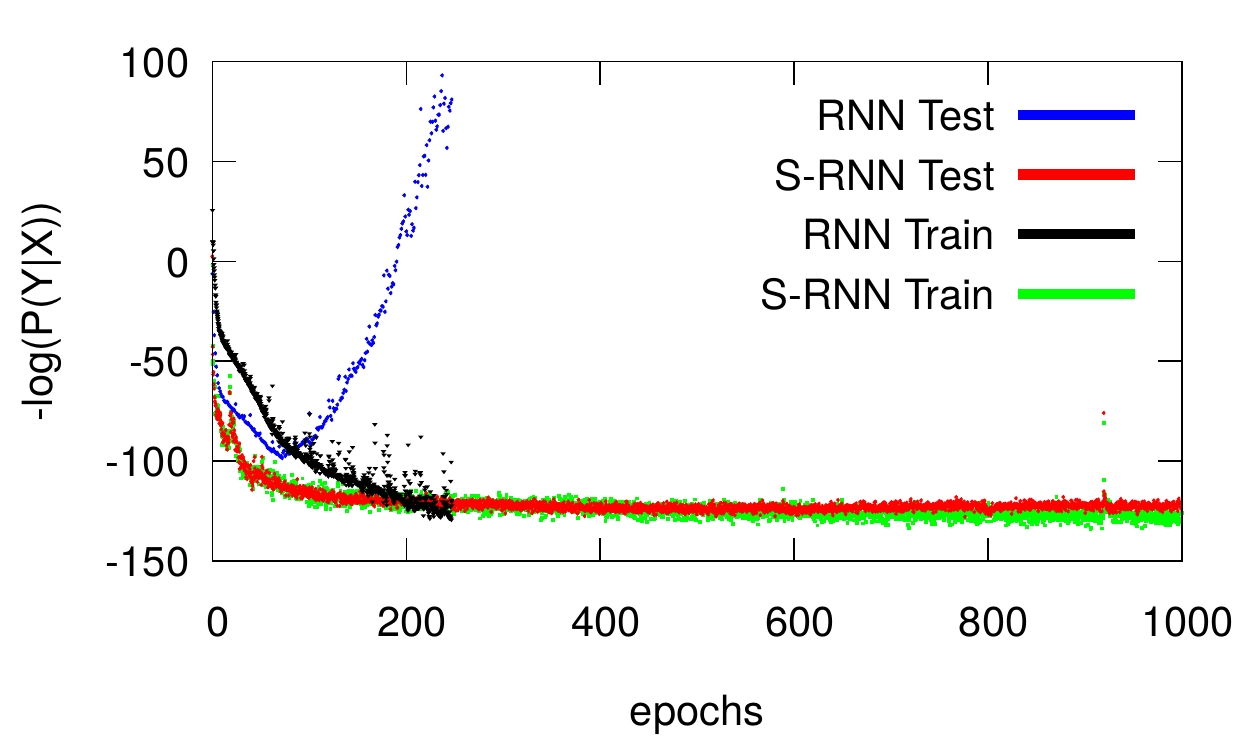} \\
    	\tiny Multivariate dynamical systems
    \end{minipage}
\caption{Learning curves on train and test sets of \strucLearn and relevant baselines.}
\label{fig:overfittingAll}
\end{figure}

\paragraph{Randomization and overfitting} An important part of \strucLearn is the randomized sampling procedure 
we used to generate the serialisations of a given complex structure. Our experimental results, which we will very briefly 
discuss, show that this offers a significant protection from overfitting.  We generated the learning curves, on both the 
training and testing sets, as a function of the training epochs for the different structures with which we experimented (figure~\ref{fig:overfittingAll}).
In the experiments with sets we see that the test performance 
follows a similar evolution to that on the training set and they never diverge. 
%
A similar picture also appears in the tree problems. 
In the graph problem we plot the learning curves on the canonical (non-randomized) and non-canonical (randomized) SMILES.
We see that in the randomized SMILES the learning curves on the training and test exhibit 
very similar relative behaviors. 
This is not the case for the canonical 
SMILES where very early the train and test learning curves diverge.  Without randomization, 
the model may overfit on a particular SMILES string. With randomization the additional complexity 
of all possible equivalent SMILES strings forces the model to generate a representation which is 
compatible with the randomization procedure. In the case of the dynamical systems problem we 
compare the learning curves of \strucLearn with a multivariate RNN. As we can see there is no 
divergence between the performance on the training and the testing set for \strucLearn. When it 
comes to \baseLearn the overfitting is very severe and happens rather early. We give
in the appendix extensive results on the learning behavior of \strucLearn on the different structures.

\vspace{-0.3cm}
\section{Conclusion}
\vspace{-0.3cm}
We have presented \strucLearn, a generic framework for performing density estimation over arbitrary complex data structures.  \strucLearn
achieves a performance on a number of different structures that is comparable or better than the performance of methods
specifically developed for these structures. Our genericity hinges on the existence of a serialization/de-serialization 
procedure for the data structures in question.  The knowledge and control of the serialization operations allow us to
transfer the structural properties of the original data onto generic 
sequences over which an RNN will learn.  The combinatorial nature of the serialization provides us with  
a potentially unlimited set of training samples. We show empirically that the diversity and uniqueness of such samples 
provides strong protection against overfitting. In fact the combinatorial nature of our serialization renders the regulariser 
that we have introduced useless, since the multiple serialisations force the learned model to factorise states in a similar 
manner to the regulariser.

\note[removed]{
\section{Discussions}

We have given, in the previous section, an algorithm which is able to learn the density estimation on an arbitrary structure if we are provided with a serializer. We have also given, a method to impose a probabilistic structure based on the notion of dependencies between different elementary component of the structure.

We have also provided a regularizer, which allows to enforce at training the domain knowledge on the probability structure. Theoretically, the regularizer is not needed if the model capacity is large enough. The regularizer is needed if we want to impose the structure in the encoder for an encoder decoder architecture.

Note that computing the constraints matrix as explained in section \ref{Regularizer} can be expensive for some structures, as it consist to find all states that are the same. For some structure the comparison can be expensive. The full pairwise comparison can be avoided by using standard data structures.

One the other hand, the non regularized version is generally cheaper as it only need to create new equivalent serialisation in a consistent (respecting equation \ref{equivalence} or \ref{Prob:state}) manner without need of comparison of states.

\section{Experiments}

We demonstrate our approach with different datasets and experiments.
All experiments were trained with a stacked-LSTM using the Adam \citep{DBLP:journals/corr/KingmaB14} optimizer.
The probability family used is the mixture or Gaussian.
 The regularizer is applied on the output of the last LSTM layer. The serialisations used for training were generated on the fly at every training iterations.

We experimented with two different settings of dynamical systems. The first setting consists of a multi-variate time-series datasets.
The second setting contain in addition some input features describing the dynamical system and its initial conditions. We denote the input feature by $x$ and the output feature (the times-series) by $y$.

Let first describe our algorithm for the problem with input features.
We define the dictionary $\mathbb{B}$ by three different operations:
\begin{equation}
\mathbb{B}=\{\operatorname{AdvanceTime}(),\operatorname{AddTS}(k,v),\operatorname{AddFeature}(k,v)\}
\end{equation}
Where $k$ is the name of an attribute and $v$ a real number. The operator $\operatorname{AdvanceTime}$ represents the advance in time of the time-series, the operator $\operatorname{AddTS}$ represents the addition of the value of one variable of the multi-variate time-series and $\operatorname{AddFeature}$ represent the addition of one of the input features.
Details on the state representation is given in section \ref{Exp_Details} of the appendix.
 The dictionary ($\mathbb{B}$) is then embedded into a two dimensional vector space with first dimension the category (combination of the operator and key) and second dimension the value (0 if it doesn't exist).
The original data can be serialized by multiples sequences depending in which order the different variables are presented and depending on the interleaving of input features and time-series informations.
For the case without supervision, we remove the $\operatorname{AddFeature}$ operator.

We use uniform sampling, except for supervised cases where we give more weight to serialisations where all input features are given first. This is done by sampling uniformly 50\% of the time and sampling with input features given first the rest of the time.
 The reason to put the $\operatorname{AddFeature}$ operator first a significant amount of the time is that to sample a conditional distribution with an RNN, the condition variable should serve as initial sequence, so we give more importance to these serializations as it is the setting on which the model will be evaluated on.
 
We name our algorithm StructRNN followed, when the regularizer is used, by a number indicating the $\lambda$ parameter.
As baseline, we use for the unsupervised case an LSTM with same hyper-parameters than our model except that it output a sequence of vector of dimension equal to the multi-variate time-series. We call this model RNN. In the supervised case, we use the encoder-decoder framework by using first a LSTM which take the input as a single vector sequence and share the last hidden-state and cell-state of the encoder to the first hidden-state and cell-state of the decoder\cite{NIPS2014_5346}.

We did three different experiments. The first is a simulated dynamical system on which a numerical solver is known. The second and third experiments are real Gait datasets of clinical interest.

\subsection{Simulated Dynamical system}

We experimented on a 3 dimensional dynamical system with a 9 dimensional input features space. We generated the ground truth by using a state of the art ordinary differential equation solver.
Details on the generation of the data is found in section \ref{DataGeneration} of the appendix.
 The number of training instances is 1000 and the length of the time-series 21.

The log likelihood on the test set is given in figure \ref{fig:SupervisedLikelihood} and the numerical results in table \ref{tab:supervised_Dyn}. We remark that our method does not over-fit i.e. the test error does not increase, contrary to our baseline the standard RNN. In fact, the train and test error of our method is very similar. This is one of the main advantage of our method a strong resistance to over-fitting. This property come from the fact that the model is forced to model the structure of the data by itself as it is the only way it can model a combinatorial amount of different serialisations. The number of possible serialisations is too huge to be completely explored in training.
In this experiment, we did not find advantage  in the use of the regularizer.
Comparison of predictions are shown in figure \ref{fig:Sample_Dyn}.

\begin{figure}
	\includegraphics[width=1\linewidth]{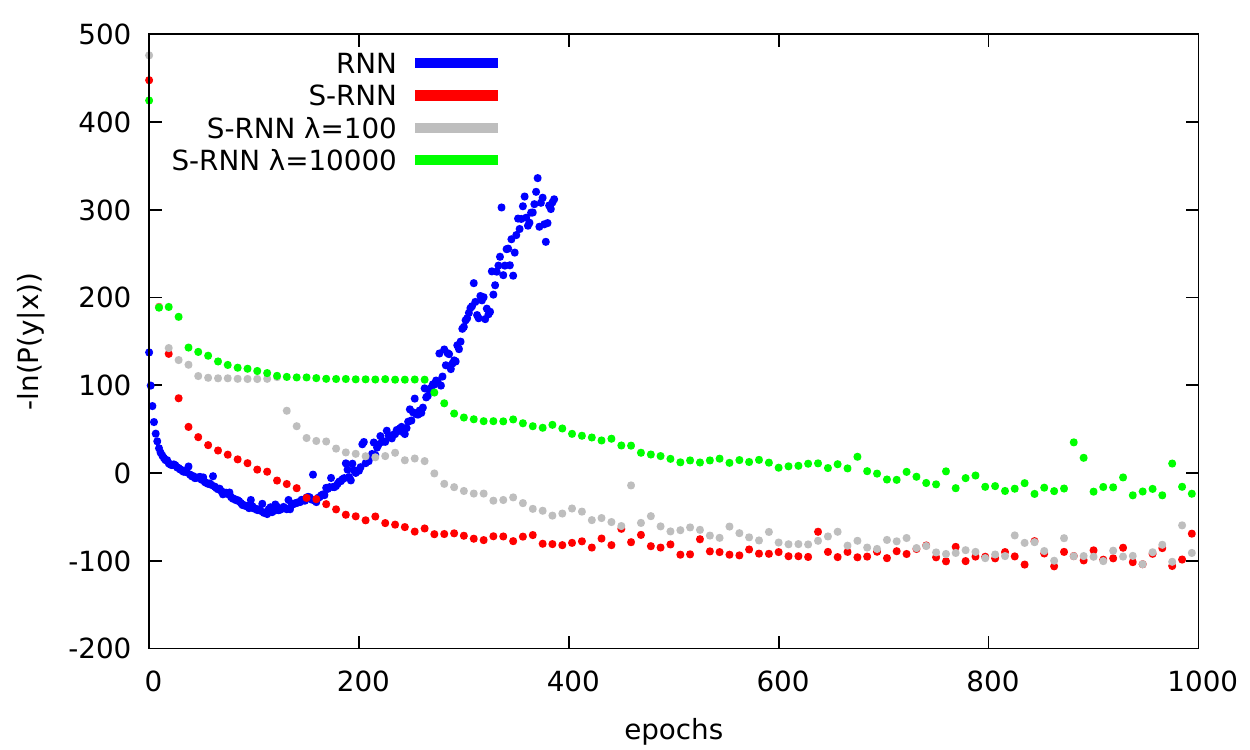}
	\caption{Plot of negative conditional log likelihood on the test set for the supervised Dynamical system problem.}
	\label{fig:SupervisedLikelihood}
\end{figure}

\begin{figure}
	\begin{minipage}{0.32\columnwidth}
		\includegraphics[width=1\linewidth,trim={2cm 0 3.1cm 0},clip]{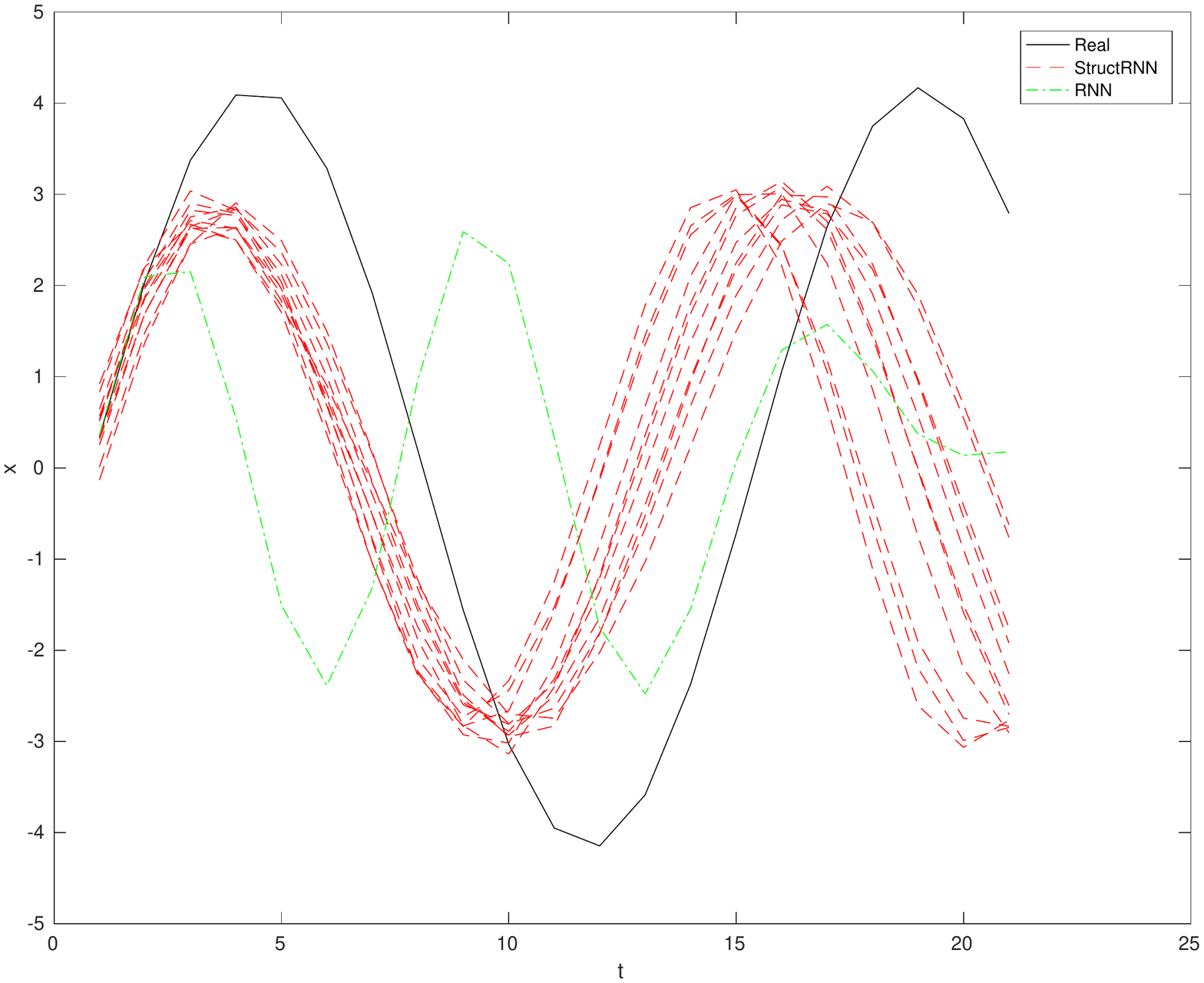}
	\end{minipage}
	\begin{minipage}{0.32\columnwidth}
	\includegraphics[width=1\linewidth,trim={2cm 0 3.1cm 0},clip]{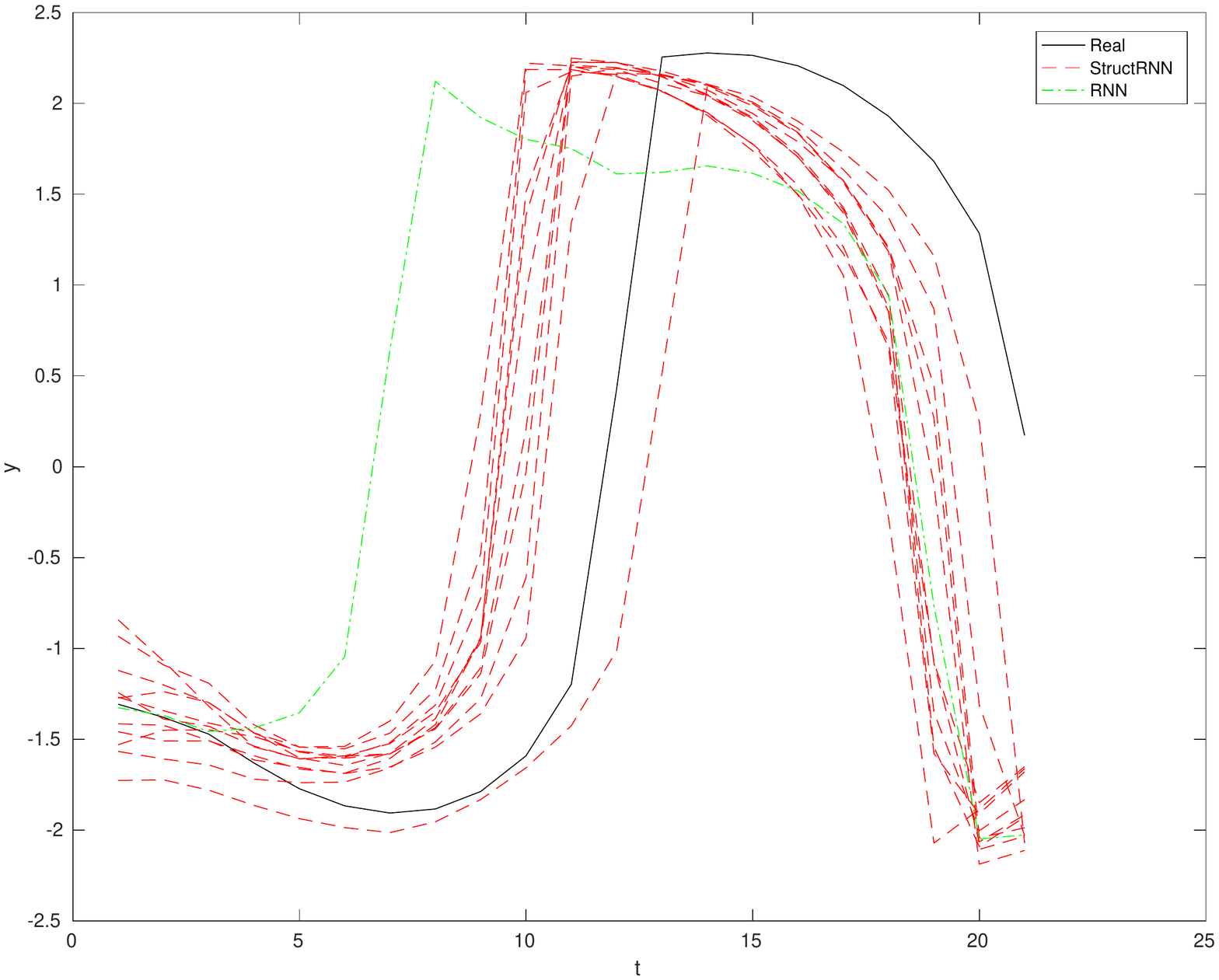}
\end{minipage}
	\begin{minipage}{0.32\columnwidth}
	\includegraphics[width=1\linewidth,trim={2cm 0 3.1cm 0},clip]{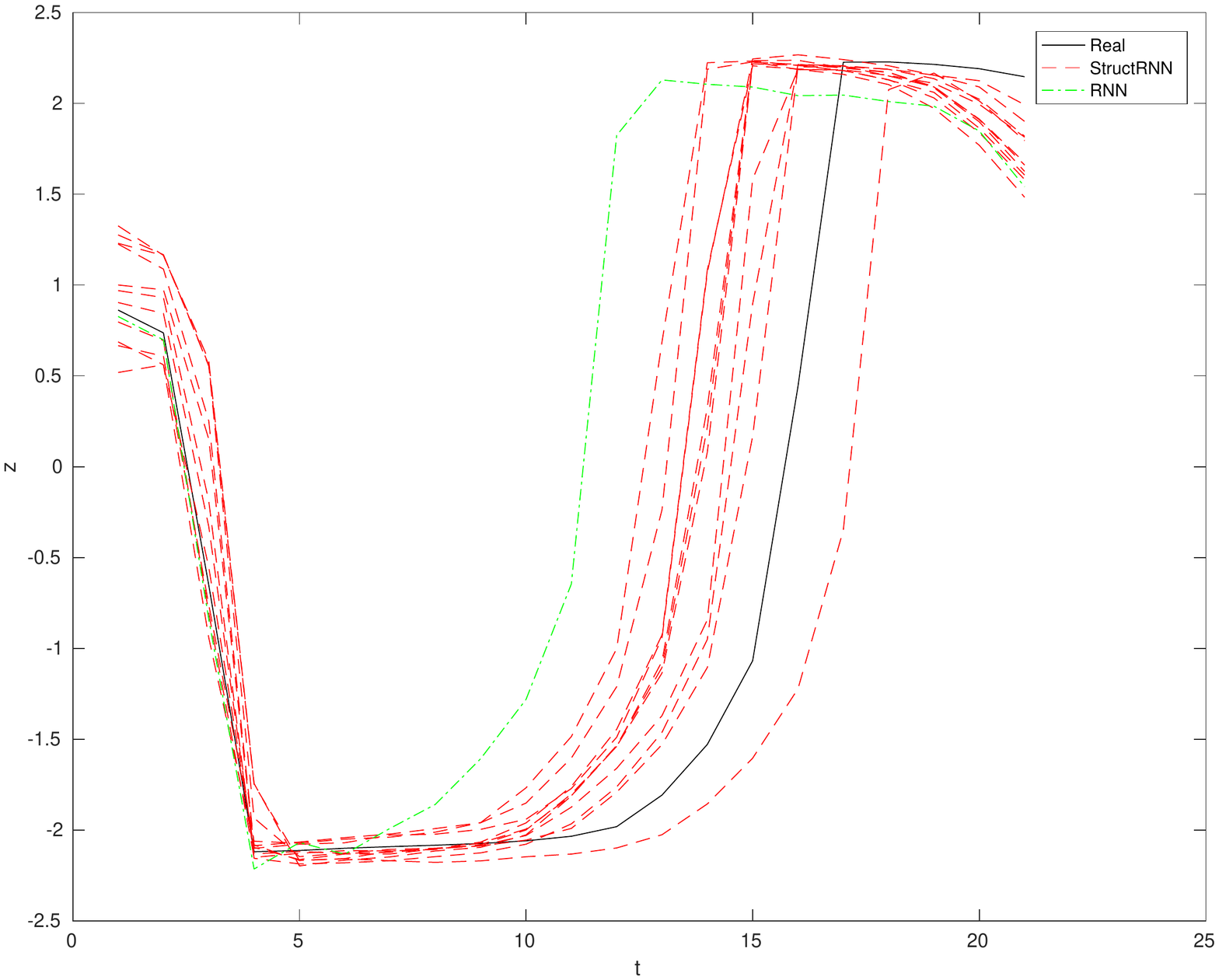}
\end{minipage}
	\caption{Comparison of predictions (most probable reconstruction) for a single instance for the simulated Dynamical system problem. The multiple curves from StructRNN come from the randomisation of the serialisation.}
	\label{fig:Sample_Dyn}
\end{figure}

	\begin{table}
	{
		\tiny
		\begin{tabular}{|c|c|c|c|c|c|c|c|}
			\hline
			StructRNN&StructRNN 100&StructRNN 10000&RNN\\
			\hline
			$\mathbf{-106}$&$\mathbf{-104}$&-57 &-40\\
			\hline
		\end{tabular}
	}
	\caption{Test negative log likelihood on the supervised Dynamical system problem. Bold indicate the lowest significatif negative log likelihood.}
	\label{tab:supervised_Dyn}
\end{table}

\subsection{Supervised-Gait}

To check the effectiveness of the regularizer on a real dataset, we applied our method on a real gait dataset, where the goal is to learn the conditional probability distribution of gait given some clinical parameters. The total number of instances is 3276 and the length of the time-series 34. The dimension of input feature is 212 with missing data and we used different number of time-series representing different gait angles. To help the model to learn the concept of missing data, we modified the sampling strategy by dropping randomly with an uniform probability the input features 50\% of the time. This allows the model to understand which input features are important without changing the conditional distribution.

As the baseline cannot treat missing data, we completed the data by the mean.
We compute the conditional log likelihood on the test set in table \ref{tab:supervised}. In this case, the regularizer bring a slight improvement. The difference in performance with the baseline is small for a single output time-series, we believe the reason of the small improvement is that most of the representative power is used to express correlation between input features, which for this setting is not useful. However, for two angles we significantly beat the baseline because we believe of a better understanding of the correlation between the two angles. We see in figure \ref{fig:overfitting} that our method did not over-fit having the same training likelihood than testing likelihood during all the training.
 To visualize the invariance property over different serializations, we visualize the effect of multiple serialisations of the clinical parameter for the most probable gait. As we see in figure \ref{fig:MultState} the differences are small and follow natural variations of gait data.
 Comparison of predictions are shown in figure \ref{fig:SampleSupervised}.
 
 	\begin{table}
 	{
 		\tiny
 		\begin{tabular}{|c|c|c|c|c|c|c|c|}
 			\hline
 			Angles&StructRNN&\multicolumn{2}{|c|}{StructRNN 100}&\multicolumn{2}{|c|}{StructRNN 10000}&\multicolumn{2}{|c|}{RNN}\\
 			\hline
 			&mean&mean&p-value&mean&p-value&mean&p-value\\
 			\hline
 			1&-47&-44&$\mathbf{2\cdot 10^{-5}}$&-49&1.0&$\mathbf{-53}$&1\\
 			2&$\mathbf{-125}$&-114&$\mathbf{0}$&-100&$\mathbf{0}$&-97&$\mathbf{0}$\\	
 			\hline
 		\end{tabular}
 	}
 	\caption{Test negative log likelihood on the supervised	 Gait problem. Bold indicate the lowest significatif negative log likelihood. A $t$-test on whether StructRNN is better is performed. }
 	\label{tab:supervised}
 \end{table}
 
 \begin{figure}
 	\includegraphics[width=1\linewidth]{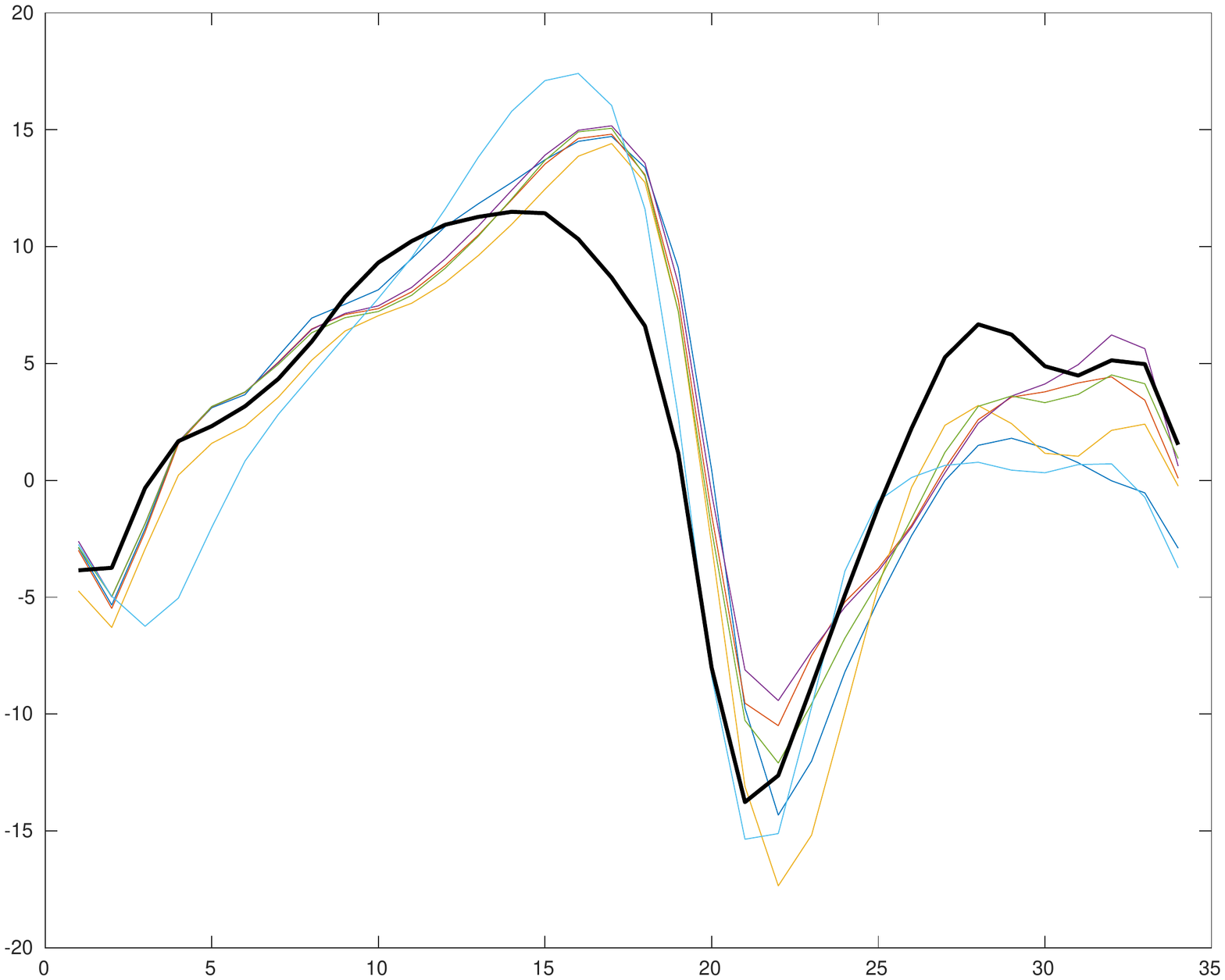}
 	\caption{Sampling of the most probable gait given equivalent order of serialization for the input feature. In thick black line is the real gait.}
 	\label{fig:MultState}
 \end{figure}
 
 \begin{figure}
 	\includegraphics[width=1\linewidth]{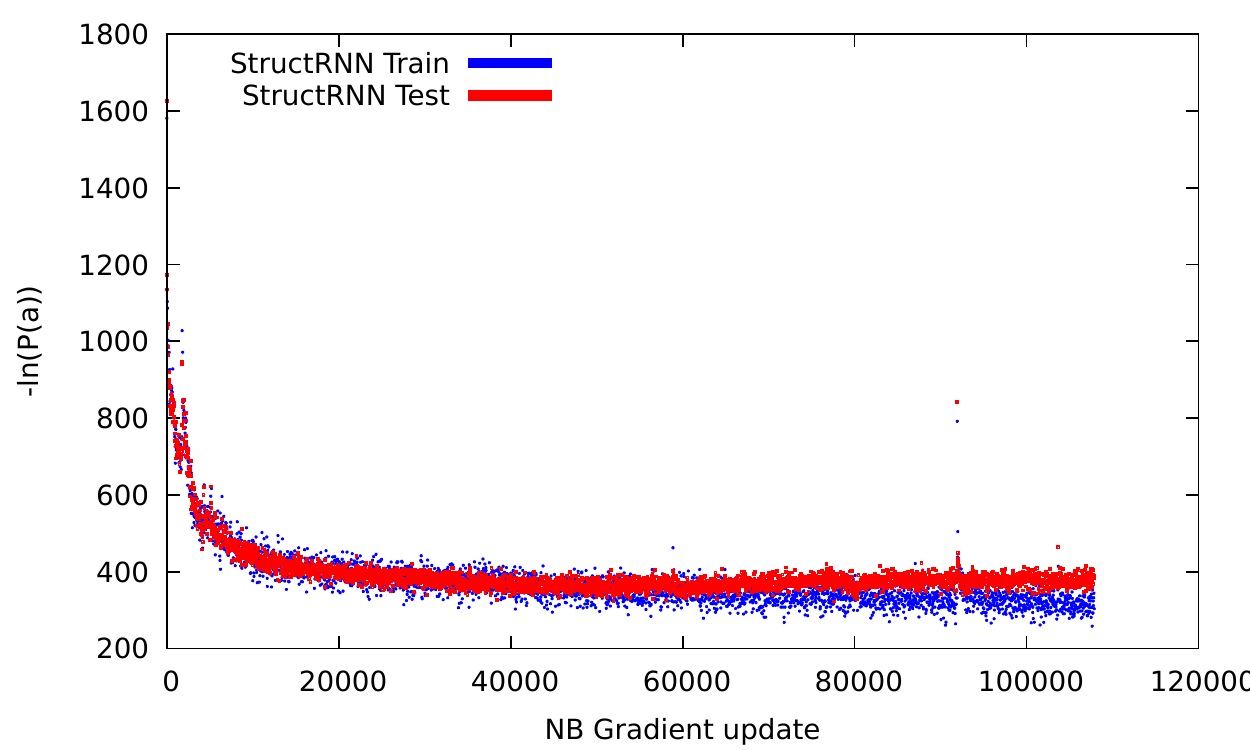}
 	\caption{Training and testing negative log likelihood of serializations for our method for the supervised gait problem.}
 	\label{fig:overfitting}
 \end{figure}

\begin{figure}
	\includegraphics[width=1\linewidth]{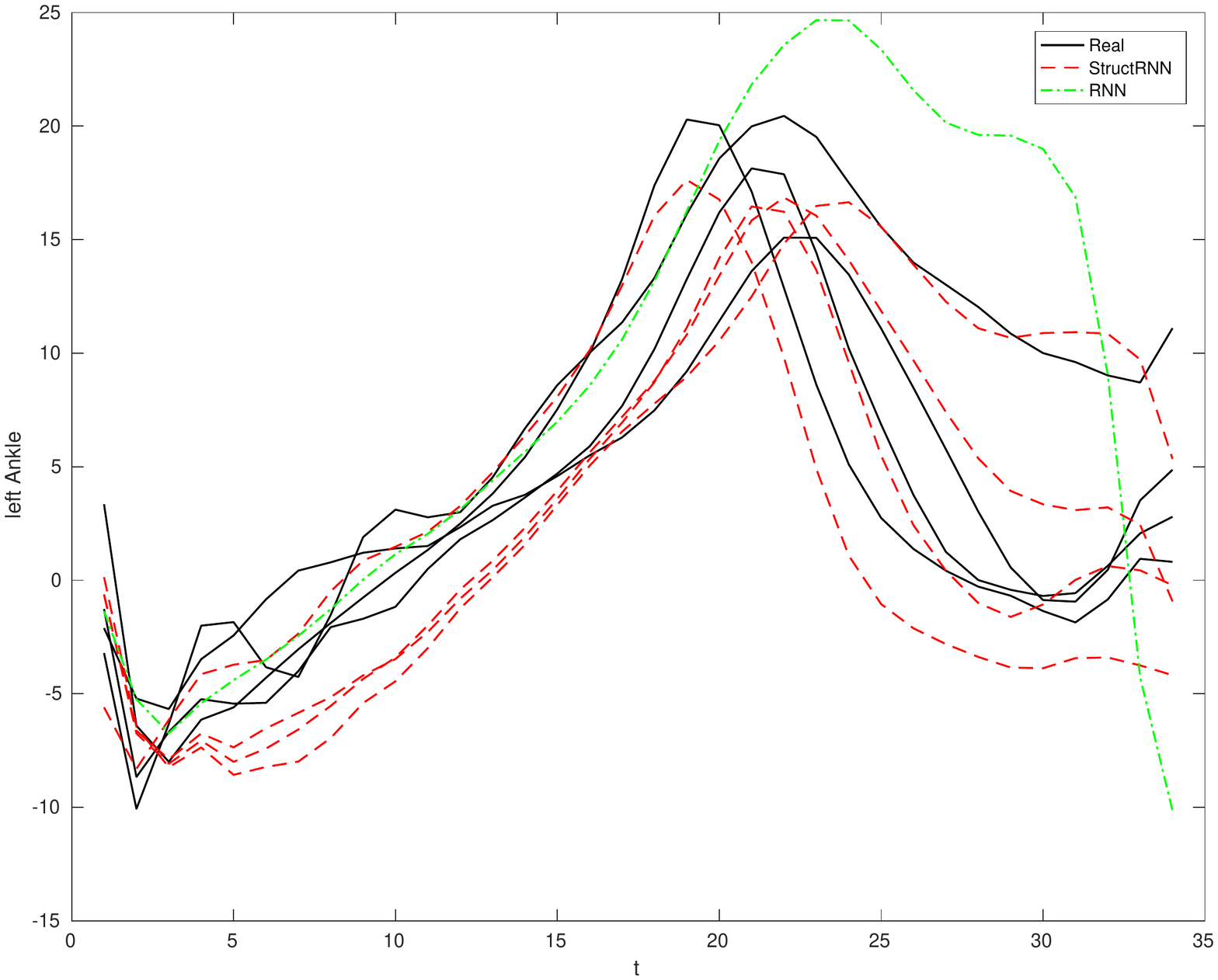}
	\caption{Comparison of predictions for a single instance for the supervised Gait problem.}
	\label{fig:SampleSupervised}
\end{figure}

\subsection{Unsupervised-Gait}
To finish, we show the test likelihood in the unsupervised case, where we ignore the input features. Numerical results are found in table \ref{tab:unsupervised}. Samples of our model is found in figure \ref{fig:SamplesStructRNN} which look similar to sample from the ground truth shown in figure \ref{fig:RealUnsupervised}.


	\begin{table}
		{
			\tiny
			\begin{center}
			\begin{tabular}{|c|c|c|}
				\hline
				Nb of Angles&StructRNN&RNN\\
				\hline
				&mean&mean\\
				\hline
				2&$\mathbf{-6.8}$&7.3\\
				4&$\mathbf{-66}$&-14\\
				8&$\mathbf{-32}$&83.3\\	
				\hline
			\end{tabular}
			\end{center}
		}
		\caption{Test negative log likelihood on the unsupervised Gait problem. Bold indicate the lowest significatif negative log likelihood.}
		\label{tab:unsupervised}
	\end{table}

\begin{table}
	\begin{tabular}{c|c}
		Problem&Number of possibility\\
		\hline
		Simulated dynamical system&$8\cdot 10^{21}$\\
		Supervised Gait&$>8\cdot 10^{300}$\\
		Unsupervised Gait 2 angles&$2\cdot 10^{6}$\\
		Unsupervised Gait 4 angles&$10^{28}$\\
		Unsupervised Gait 8 angles&$5\cdot10^{96}$\\
	\end{tabular}
\caption{Number of possible serialisations per instances}
\label{fig:nb_Serialisation}
\end{table}

\begin{figure}
	\begin{minipage}{.45\columnwidth}
	\centering
	\includegraphics[width=1\columnwidth]{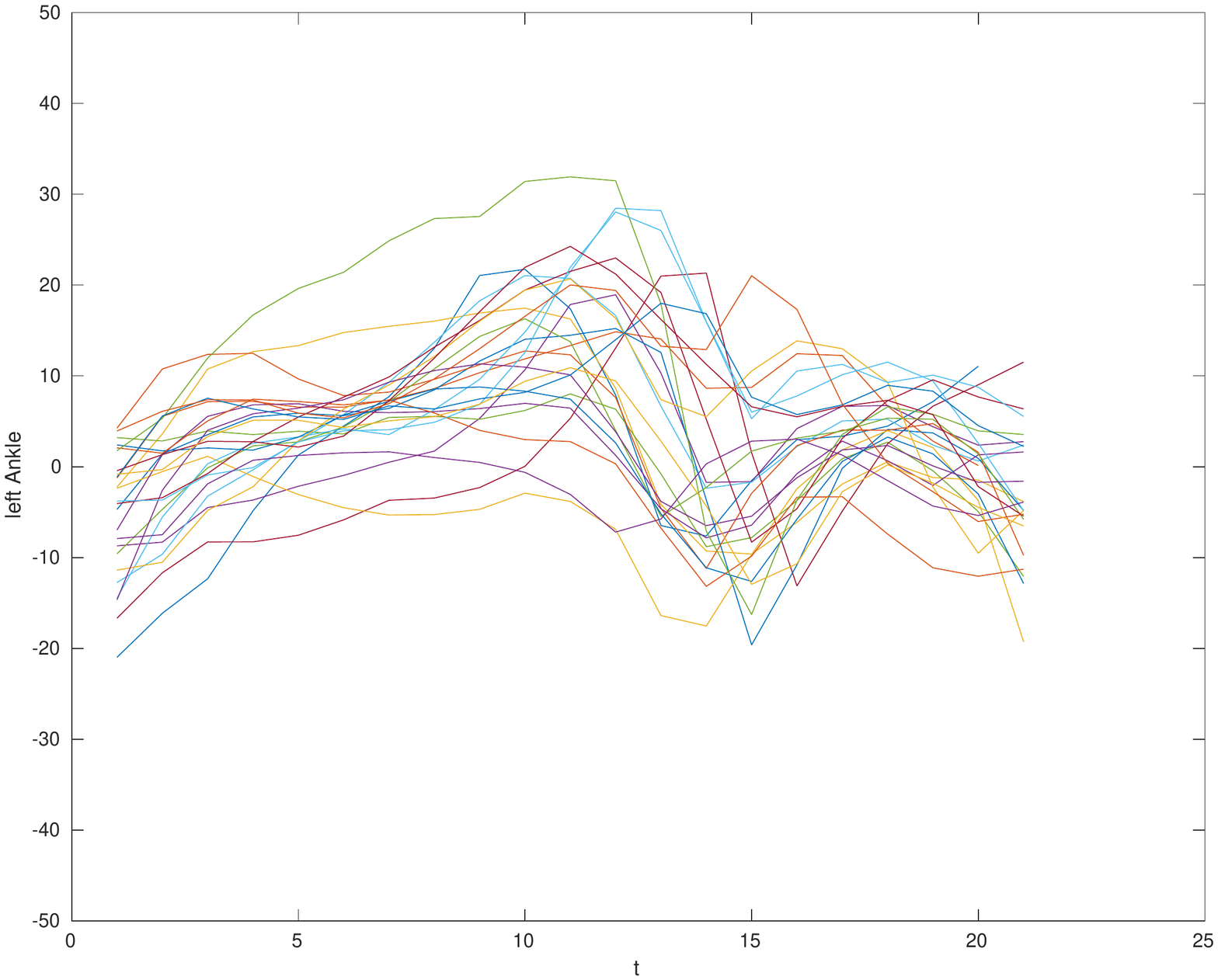}
	\caption{Samples from StrucRNN for the unsupervised Gait problem.}
		\label{fig:SamplesStructRNN}
		\end{minipage}\hfil\begin{minipage}{.45\columnwidth}
		\centering
		\includegraphics[width=1\columnwidth]{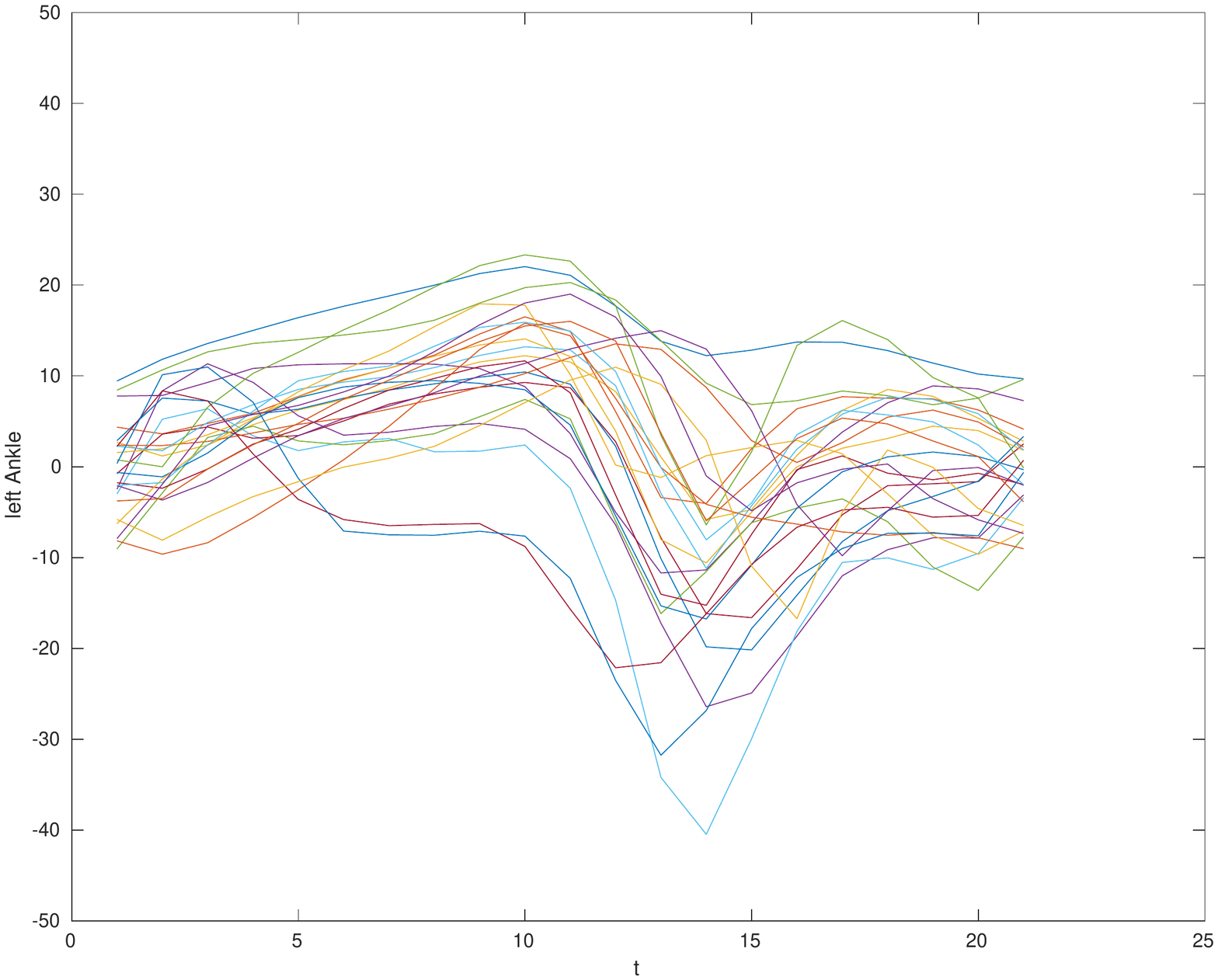}
		\caption{Real samples for the unsupervised Gait problem.}
		\label{fig:RealUnsupervised}
		\end{minipage}
	\centering
	\begin{minipage}{.45\columnwidth}
	\includegraphics[width=1\columnwidth]{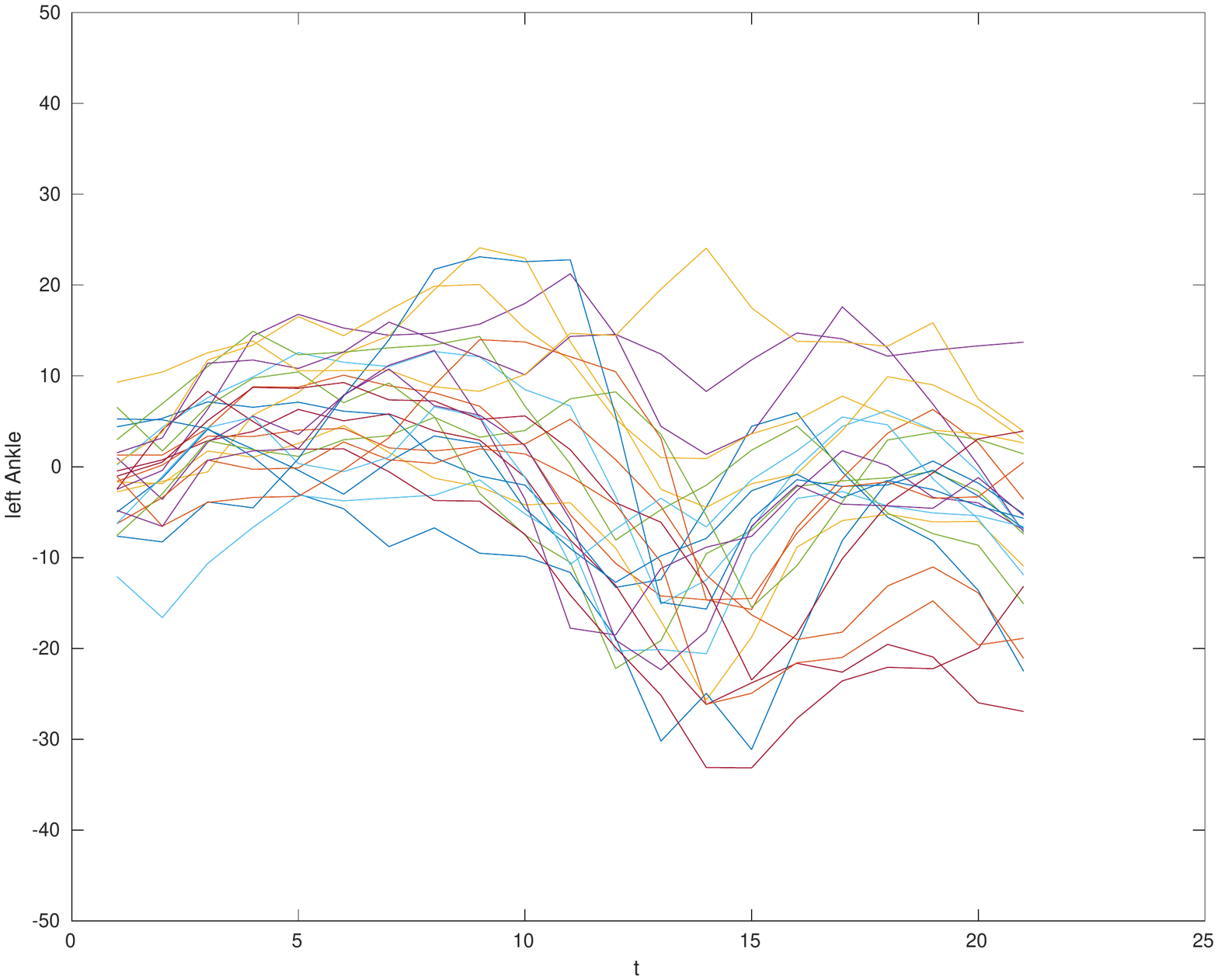}
	\caption{Samples from RNN for the unsupervised Gait problem.}
	\label{fig:SamplesRNN}
\end{minipage}
\end{figure}

\subsection{Results discussions}

One interesting property implementation-wise is that the regularizer does not bring improvement. This mean that sampling serialisation in a structure aware manner bring the same information than the regularizer. This property can be seen in figure \ref{fig:MultState} where the most probable sample condition on multiple equivalent serialisations are very similar. As the regularizer is expensive to compute in general, this property make the problem cheaper.

Another interesting property of our method is resistance to over-fitting this is an interesting property as it allows to improve the performance of the algorithm simply by increasing the capacity of the model. This resistance to over-fitting come from the multiplicity of equivalent serialization(see table \ref{fig:nb_Serialisation} for the number of equivalent serialisation per instance for all our experiments) which has two effects on the model. First simply memorising the serialisations is not possible as there are too many of them. Instead, the model is forced to build a hidden state representation which respect the structure. To build this representation, it will use links and correlation between variables. Learning to respect the constraint make our algorithm slower at start as it need to understand the concept however the speed of improvement is then similar to our baseline.

A last interesting property, is that because each element of the sequence live in a smaller dimensional space, the probability of the next element given the  past is simpler. However this is done at the cost of a longer sequence. In another word, we trade-off simplicity of the probability knowing the state with difficulty to learn the state.

\section{Conclusion}

We have shown in this paper, how we can learn over complex structure by serializing the structure into a sequence and train a standard sequence density estimator on the serialization.
We have also shown that a clever serialization allows to model invariance.
Experiments show that the multiple possible serialization provide a resistance against overfiting.
}

\bibliography{main}

\begin{thebibliography}{10}

\bibitem{alvarez-melis_tree-structured_2016}
David Alvarez-Melis and Tommi~S. Jaakkola.
\newblock Tree-structured decoding with doubly-recurrent neural networks.
\newblock November 2016.

\bibitem{PixelRNN}
Maria{-}Florina Balcan and Kilian~Q. Weinberger, editors.
\newblock {\em Proceedings of the 33nd International Conference on Machine
  Learning, {ICML} 2016, New York City, NY, USA, June 19-24, 2016}, volume~48
  of {\em {JMLR} Workshop and Conference Proceedings}. JMLR.org, 2016.

\bibitem{battaglia_relational_2018}
Peter~W. Battaglia, Jessica~B. Hamrick, Victor Bapst, Alvaro Sanchez-Gonzalez,
  Vinicius Zambaldi, Mateusz Malinowski, Andrea Tacchetti, David Raposo, Adam
  Santoro, Ryan Faulkner, Caglar Gulcehre, Francis Song, Andrew Ballard, Justin
  Gilmer, George Dahl, Ashish Vaswani, Kelsey Allen, Charles Nash, Victoria
  Langston, Chris Dyer, Nicolas Heess, Daan Wierstra, Pushmeet Kohli, Matt
  Botvinick, Oriol Vinyals, Yujia Li, and Razvan Pascanu.
\newblock Relational inductive biases, deep learning, and graph networks.
\newblock {\em arXiv:1806.01261 [cs, stat]}, June 2018.
\newblock arXiv: 1806.01261.

\bibitem{bengio_curriculum_nodate}
Yoshua Bengio, Jérôme Louradour, Ronan Collobert, and Jason Weston.
\newblock {\em Curriculum {Learning}}.

\bibitem{BoostLibrary}
Boost.
\newblock {Boost C++ Libraries}.
\newblock \url{http://www.boost.org/}, 2019.

\bibitem{brown_guacamol:_2018}
Nathan Brown, Marco Fiscato, Marwin H.~S. Segler, and Alain~C. Vaucher.
\newblock {GuacaMol}: {Benchmarking} {Models} for {De} {Novo} {Molecular}
  {Design}.
\newblock {\em arXiv:1811.09621 [physics, q-bio]}, November 2018.
\newblock arXiv: 1811.09621.

\bibitem{cho_learning_2014}
Kyunghyun Cho, Bart van Merrienboer, Caglar Gulcehre, Dzmitry Bahdanau, Fethi
  Bougares, Holger Schwenk, and Yoshua Bengio.
\newblock Learning {Phrase} {Representations} using {RNN} {Encoder}-{Decoder}
  for {Statistical} {Machine} {Translation}.
\newblock {\em arXiv:1406.1078 [cs, stat]}, June 2014.
\newblock arXiv: 1406.1078.

\bibitem{torch}
R.~Collobert, K.~Kavukcuoglu, and C.~Farabet.
\newblock Torch7: A matlab-like environment for machine learning.
\newblock In {\em BigLearn, NIPS Workshop}, 2011.

\bibitem{TreeDongL16}
Li~Dong and Mirella Lapata.
\newblock Language to logical form with neural attention.
\newblock In {\em Proceedings of the 54th Annual Meeting of the Association for
  Computational Linguistics, {ACL} 2016, August 7-12, 2016, Berlin, Germany,
  Volume 1: Long Papers}, 2016.

\bibitem{DBLP:journals/ijon/GallicchioM13}
Claudio Gallicchio and Alessio Micheli.
\newblock Tree echo state networks.
\newblock {\em Neurocomputing}, 101:319--337, 2013.

\bibitem{gilmer_neural_2017}
Justin Gilmer, Samuel~S. Schoenholz, Patrick~F. Riley, Oriol Vinyals, and
  George~E. Dahl.
\newblock Neural {Message} {Passing} for {Quantum} {Chemistry}.
\newblock {\em arXiv:1704.01212 [cs]}, April 2017.
\newblock arXiv: 1704.01212.

\bibitem{moleculeDuvenaud}
Rafael G{\'{o}}mez{-}Bombarelli, David~K. Duvenaud, Jos{\'{e}}~Miguel
  Hern{\'{a}}ndez{-}Lobato, Jorge Aguilera{-}Iparraguirre, Timothy~D. Hirzel,
  Ryan~P. Adams, and Al{\'{a}}n Aspuru{-}Guzik.
\newblock Automatic chemical design using a data-driven continuous
  representation of molecules.
\newblock {\em CoRR}, abs/1610.02415, 2016.

\bibitem{GAN}
Ian~J. Goodfellow, Jean Pouget{-}Abadie, Mehdi Mirza, Bing Xu, David
  Warde{-}Farley, Sherjil Ozair, Aaron~C. Courville, and Yoshua Bengio.
\newblock Generative adversarial networks.
\newblock {\em CoRR}, abs/1406.2661, 2014.

\bibitem{GravesSequence}
Alex Graves.
\newblock Generating sequences with recurrent neural networks.
\newblock {\em CoRR}, abs/1308.0850, 2013.

\bibitem{LSTM}
Sepp Hochreiter and J{\"{u}}rgen Schmidhuber.
\newblock Long short-term memory.
\newblock {\em Neural Computation}, 9(8):1735--1780, 1997.

\bibitem{c++17}
{ISO}.
\newblock {\em {ISO\slash IEC 14882:2017 Information technology --- Programming
  languages --- C++}}.
\newblock Fifth edition, December 2017.

\bibitem{kanezaki_rotationnet:_2018}
Asako Kanezaki, Yasuyuki Matsushita, and Yoshifumi Nishida.
\newblock {RotationNet}: {Joint} {Object} {Categorization} and {Pose}
  {Estimation} {Using} {Multiviews} {From} {Unsupervised} {Viewpoints}.
\newblock pages 5010--5019, 2018.

\bibitem{DBLP:journals/corr/KingmaB14}
Diederik~P. Kingma and Jimmy Ba.
\newblock Adam: {A} method for stochastic optimization.
\newblock {\em CoRR}, abs/1412.6980, 2014.

\bibitem{VAEKingmaW13}
Diederik~P. Kingma and Max Welling.
\newblock Auto-encoding variational bayes.
\newblock {\em CoRR}, abs/1312.6114, 2013.

\bibitem{kusner_grammar_2017}
Matt~J. Kusner, Brooks Paige, and Jos{\'e}~Miguel Hern{\'a}ndez-Lobato.
\newblock Grammar variational autoencoder.
\newblock In Doina Precup and Yee~Whye Teh, editors, {\em Proceedings of the
  34th International Conference on Machine Learning}, volume~70 of {\em
  Proceedings of Machine Learning Research}, pages 1945--1954, International
  Convention Centre, Sydney, Australia, 06--11 Aug 2017. PMLR.

\bibitem{rdkit}
Greg Landrum.
\newblock Rdkit: Open-source cheminformatics.

\bibitem{lavrac2001relational}
Nada Lavrac and Saso Dzeroski.
\newblock Relational data mining, 2001.

\bibitem{TreeLiuXGGLW11}
Han Liu, Min Xu, Haijie Gu, Anupam Gupta, John~D. Lafferty, and Larry~A.
  Wasserman.
\newblock Forest density estimation.
\newblock {\em Journal of Machine Learning Research}, 12:907--951, 2011.

\bibitem{RedOlivaDPXS17}
Junier~B. Oliva, Kumar~Avinava Dubey, Barnab{\'{a}}s P{\'{o}}czos, Eric~P.
  Xing, and Jeff~G. Schneider.
\newblock Recurrent estimation of distributions.
\newblock {\em CoRR}, abs/1705.10750, 2017.

\bibitem{pytorch}
Adam Paszke, Sam Gross, Soumith Chintala, Gregory Chanan, Edward Yang, Zachary
  DeVito, Zeming Lin, Alban Desmaison, Luca Antiga, and Adam Lerer.
\newblock Automatic differentiation in pytorch.
\newblock In {\em NIPS-W}, 2017.

\bibitem{Ramsundar-et-al-2019}
Bharath Ramsundar, Peter Eastman, Patrick Walters, Vijay Pande, Karl Leswing,
  and Zhenqin Wu.
\newblock {\em Deep Learning for the Life Sciences}.
\newblock O'Reilly Media, 2019.
\newblock
  \url{https://www.amazon.com/Deep-Learning-Life-Sciences-Microscopy/dp/1492039837}.

\bibitem{VAERezende}
Danilo~Jimenez Rezende, Shakir Mohamed, and Daan Wierstra.
\newblock Stochastic backpropagation and approximate inference in deep
  generative models.
\newblock In {\em Proceedings of the 31th International Conference on Machine
  Learning, {ICML} 2014, Beijing, China, 21-26 June 2014}, pages 1278--1286,
  2014.

\bibitem{PCL}
Radu~Bogdan Rusu and Steve Cousins.
\newblock {3D is here: Point Cloud Library (PCL)}.
\newblock In {\em {IEEE International Conference on Robotics and Automation
  (ICRA)}}, Shanghai, China, May 9-13 2011.

\bibitem{GeneratifVAE}
Kihyuk Sohn, Honglak Lee, and Xinchen Yan.
\newblock Learning structured output representation using deep conditional
  generative models.
\newblock In {\em Advances in Neural Information Processing Systems 28: Annual
  Conference on Neural Information Processing Systems 2015, December 7-12,
  2015, Montreal, Quebec, Canada}, pages 3483--3491, 2015.

\bibitem{song_3d_2015}
{and}~S. Song, A.~Khosla, {and}, and {and}~J. Xiao.
\newblock 3d {ShapeNets}: {A} deep representation for volumetric shapes.
\newblock In {\em 2015 {IEEE} {Conference} on {Computer} {Vision} and {Pattern}
  {Recognition} ({CVPR})}, pages 1912--1920, June 2015.

\bibitem{Seq2SeqSutskeverVL14}
Ilya Sutskever, Oriol Vinyals, and Quoc~V. Le.
\newblock Sequence to sequence learning with neural networks.
\newblock In {\em Advances in Neural Information Processing Systems 27: Annual
  Conference on Neural Information Processing Systems 2014, December 8-13 2014,
  Montreal, Quebec, Canada}, pages 3104--3112, 2014.

\bibitem{NIPS2014_5346}
Ilya Sutskever, Oriol Vinyals, and Quoc~V Le.
\newblock Sequence to sequence learning with neural networks.
\newblock In Z.~Ghahramani, M.~Welling, C.~Cortes, N.~D. Lawrence, and K.~Q.
  Weinberger, editors, {\em Advances in Neural Information Processing Systems
  27}, pages 3104--3112. Curran Associates, Inc., 2014.

\bibitem{sol2}
ThePhD.
\newblock sol2: a c++ {$<$-$>$} lua api wrapper.
\newblock [Software], Mai 2019.
\newblock https://github.com/ThePhD/sol2.

\bibitem{NadeUriaCGML16}
Benigno Uria, Marc{-}Alexandre C{\^{o}}t{\'{e}}, Karol Gregor, Iain Murray, and
  Hugo Larochelle.
\newblock Neural autoregressive distribution estimation.
\newblock {\em Journal of Machine Learning Research}, 17:205:1--205:37, 2016.

\bibitem{WaveNet}
A{\"{a}}ron van~den Oord, Sander Dieleman, Heiga Zen, Karen Simonyan, Oriol
  Vinyals, Alex Graves, Nal Kalchbrenner, Andrew~W. Senior, and Koray
  Kavukcuoglu.
\newblock Wavenet: {A} generative model for raw audio.
\newblock {\em CoRR}, abs/1609.03499, 2016.

\bibitem{SetVinyalsBK15}
Oriol Vinyals, Samy Bengio, and Manjunath Kudlur.
\newblock Order matters: Sequence to sequence for sets.
\newblock {\em CoRR}, abs/1511.06391, 2015.

\bibitem{weininger_smiles_2002}
David Weininger.
\newblock {SMILES}, a chemical language and information system. 1.
  {Introduction} to methodology and encoding rules, May 2002.

\bibitem{Deep_Sets}
Manzil Zaheer, Satwik Kottur, Siamak Ravanbakhsh, Barnabas Poczos, Ruslan~R
  Salakhutdinov, and Alexander~J Smola.
\newblock Deep sets.
\newblock In I.~Guyon, U.~V. Luxburg, S.~Bengio, H.~Wallach, R.~Fergus,
  S.~Vishwanathan, and R.~Garnett, editors, {\em Advances in Neural Information
  Processing Systems 30}, pages 3394--3404. Curran Associates, Inc., 2017.

\bibitem{TreeZhangLL16}
Xingxing Zhang, Liang Lu, and Mirella Lapata.
\newblock Top-down tree long short-term memory networks.
\newblock In {\em {NAACL} {HLT} 2016, The 2016 Conference of the North American
  Chapter of the Association for Computational Linguistics: Human Language
  Technologies, San Diego California, USA, June 12-17, 2016}, pages 310--320,
  2016.

\bibitem{TreeZhouLCXLCH17}
Ganbin Zhou, Ping Luo, Rongyu Cao, Yijun Xiao, Fen Lin, Bo~Chen, and Qing He.
\newblock Generative neural machine for tree structures.
\newblock {\em CoRR}, abs/1705.00321, 2017.

\end{thebibliography}
\bibliographystyle{plain}
\newpage
\appendix
\section{Algorithm}
As explained in section \ref{ProbStructure} the main goal of the algorithm is to sample a serialization $a$ for a given instance $x$ with importance of a serialization with respect to others given by $\mu$.
Because $\mu$ depend on the state, we need to construct the serialization element by element however we also need to ensure that the deserialization  of the constructed serialization is $x$.
Our strategy is the following:
\begin{enumerate}
	\item List all possible serializations.
	\item For every time step $t$ do:
	\begin{enumerate}
		\item Compute the set of possible next element.
		\item Sample the next element from the set of possible next element.
		\item Update the state and the list of all possible serializations.
	\end{enumerate}
\end{enumerate}
Theses steps are explained in details in our pseudo-code \ref{alg:Sampling} and \ref{alg:UtilsFunction}.
Note that in practice for common structures we do not need to compute the set of all serializations. It is only needed to know the set of next element that create a serialization of the instance $x$ we are interested on. Because a majority of common structures have a serialization algorithm which is recursive in nature this can be much faster to implement than the general algorithm.

\begin{algorithm}
	\begin{algorithmic}
		\Function{Serialisation}{$x$,$X$,$f$,$\mu$}
		\State $a_{\rm sample}\gets ()$ \Comment{Sampled serialisation from $\mathbb{A}$.}
		\State $s\gets s_0$  \Comment{Current state which is initialized to $s_0$.}
		\State $A_{\rm all}\gets X^{-1}(x)$ \Comment{Set of candidate serialisations.}
		\State $t \gets 1$
		\Repeat
		\State $\mathcal L \gets $ PossibleElement($t$,$A_{\rm all}$)\Comment{List of all possible next elements}
		\State $a^{\rm next} \gets $ Sample($\mu$,$s$,$\mathcal L$) \Comment{Sample $\mathcal L$ according to $\mu$ and $s$}
		\State $a_{\rm sample}\gets (a_{\rm sample}, a^{\rm next})$ \Comment{Concatenate the sampled element to the current serialization.}
		\State $s\gets f(s,a^{next})$\Comment{Update the current state.}
		\State $A_{\rm all} \gets $ UpdateList($A_{\rm all}$,$a^{\rm next}$,$t$) \Comment{Remove from $A_{\rm all}$ serialisations not having $a^{\rm next}$ at position $t$. }
		\State $t\gets t+1$
		\Until{$a^{\rm next}=$ eos} \Comment{Stop at the end of the sequence.}
		\State \textbf{return} $a$
		\EndFunction
	\end{algorithmic}
	\caption{Sampling serializations}
	\label{alg:Sampling}
\end{algorithm}

\begin{algorithm}
	\begin{algorithmic}
		\Function{PossibleElement}{$t$,$A_{\rm all}$}
		\State $\mathcal L\gets\{\}$
		\ForAll{$a\in A_{\rm all}$}
		\State $\mathcal L\gets \mathcal L\cup \{a^{t}\}$ \Comment{Add to $\mathcal L$ element at position $t$ of the sequence $a$.}
		\EndFor
		\State \textbf{return} $\mathcal L$
		\EndFunction
		\State
		\Function{Sample}{$\mu$,$s$,$\mathcal L$}
		\State ${\rm norm}\gets\sum_{a\in \mathcal L} \mu(a,s)$\Comment{Compute the total weight on $a$ for given $s$.}
		\ForAll {$a\in \mathcal L $}
		\State $P_{a}\gets \frac{ \mu(a,s)}{\rm norm}$ \Comment{Create a probability distribution on $a$.}
		\EndFor
		\State $a$ is sampled with probability distribution $P$.
		\State \textbf{return} $a$
		\EndFunction
		\State
		\Function{UpdateList}{$A_{\rm all}$,$a^{\rm next}$,$t$}
		\State ${A}_{\rm temp}\gets \{\}$
		\ForAll {$a\in A_{\rm all} $}
		\If{$a^{t}=a^{\rm next}$}\Comment{Check if the $t$ element is $a^{\rm next}$.}
		\State ${A}_{\rm temp}\gets {A}_{\rm temp}\cup \{a\}$\Comment{If it is the case keep the whole sequence.}
		\EndIf
		\EndFor
		\State \textbf{return} ${A}_{\rm temp}$
		\EndFunction
	\end{algorithmic}
	\caption{Auxiliary functions.}
	\label{alg:UtilsFunction}
\end{algorithm}

\section{Detailed descriptions of experiments and results}
\subsection{Set problems}
\label{Long-Set}

We use \strucLearn to solve a classification task over sets. The task is to classify the 3d model of a shape represented as a set of points. We experiment with two datasets
 ModelNet10 and ModelNet40, \citep{song_3d_2015}, which respectively contain 10 and 40 objects types, such as airplane, xbox, stair, car, \ldots. Each object has a minimum 
of 100 different instances. We use the official split into train and test sets. The total number of instances is 4896 and 12312 respectively.

We normalize the data into the unit cube using an homogeneous scale and a translation. We represent the points with a six-dimensional vector 
given by their $x,y,z$ coordinates and their squares. 
We added the square as feature because it allows to convert the data into cylindrical and spherical coordinate 
by a linear transformation.  We believe that some important features are easier to compute in cylindrical/spherical coordinate. 
In addition, we added a learned feature extractor as an MLP.

To sample the points of a given object, we sample uniformly points on the surface of its model.
This is done by first representing the 3d model as a set of triangles. We then compute the surface 
of the 3d model by summing the contribution of every triangle. We then sample points by first sampling 
a triangle proportionally to its surface and then uniformly sampling a point on the surface of the triangle.

The resulting serialization strategy is a sequence of the sampled points.
Note that the number of possible serializations is infinite when neglecting 
floating point precision as there is an infinite amount of points that can be sampled on a triangle.

After tuning on the validation set build by using 100 instances we removed from the training set from every class, 
we set to the following model architecture. The tuning of the network hyper-parameters was done in the set of $[16,32,64,128,256]$. 
\note[Alex-NIPS19]{This does not look like we have set for a $\lambda$.}
As feature extractor, we use a 64 unit MLP with a single hidden layer transforming the 6 dimensional feature to a 64 dimensional feature which will then be input to the GRU. 
\note[Alex-NIPS19]{I do not understand what this preprocessing does, on what does it operate? on the instances?}
To learn of the serialisation we use a GRU\citep{cho_learning_2014} 
with 64 units over which we add a single layer with 64 hidden units to do the classification.
As optimizer we use Adam\citep{DBLP:journals/corr/KingmaB14} with a learning rate of $10^{-4}$ together with gradient clipping to the range of $[-5,5]$.
We also experimented with curriculum learning\citep{bengio_curriculum_nodate} by starting with a length of 10 samples and increasing by 4 the number of sample every epoch.
This did not change the result as can be seen in table \ref{tab:PointCloudResult}.

\begin{figure}[htbp]
	\centering
	\begin{minipage}{0.4\linewidth}
		\includegraphics[width=\linewidth]{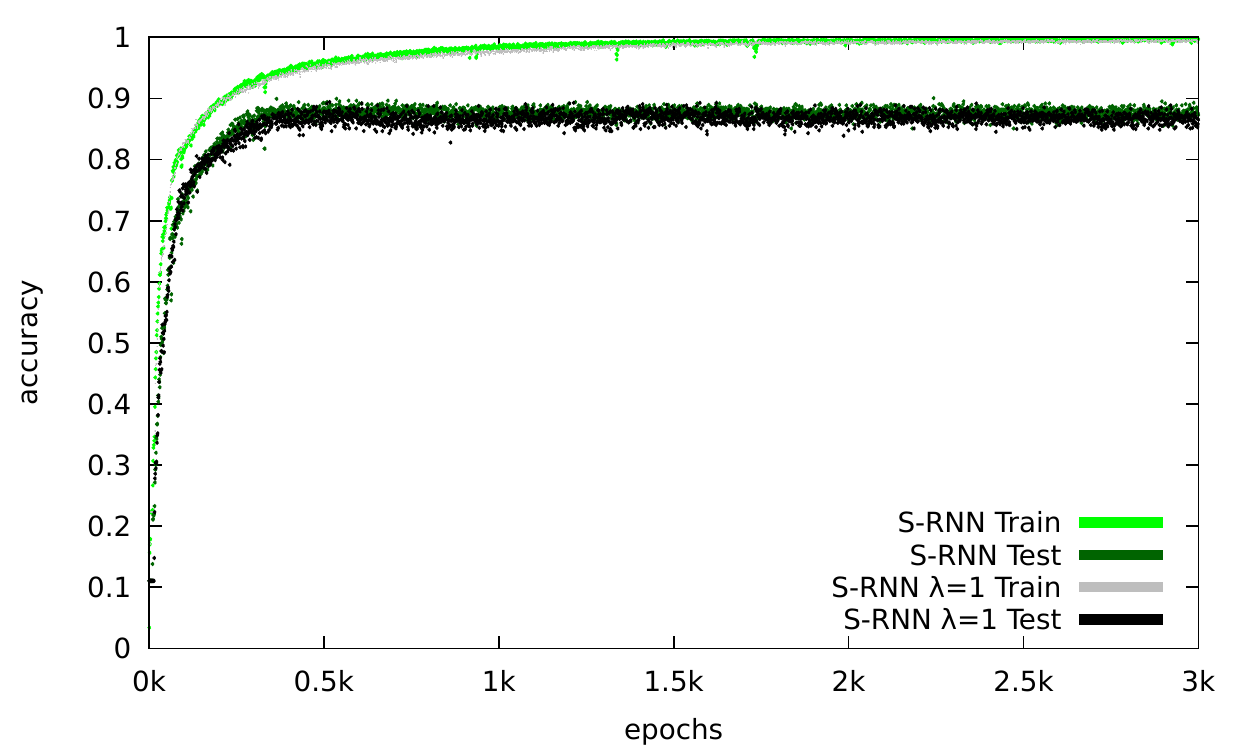}
	        \centering ModelNet10 
	\end{minipage}
	\begin{minipage}{0.4\linewidth}
	        \includegraphics[width=\linewidth]{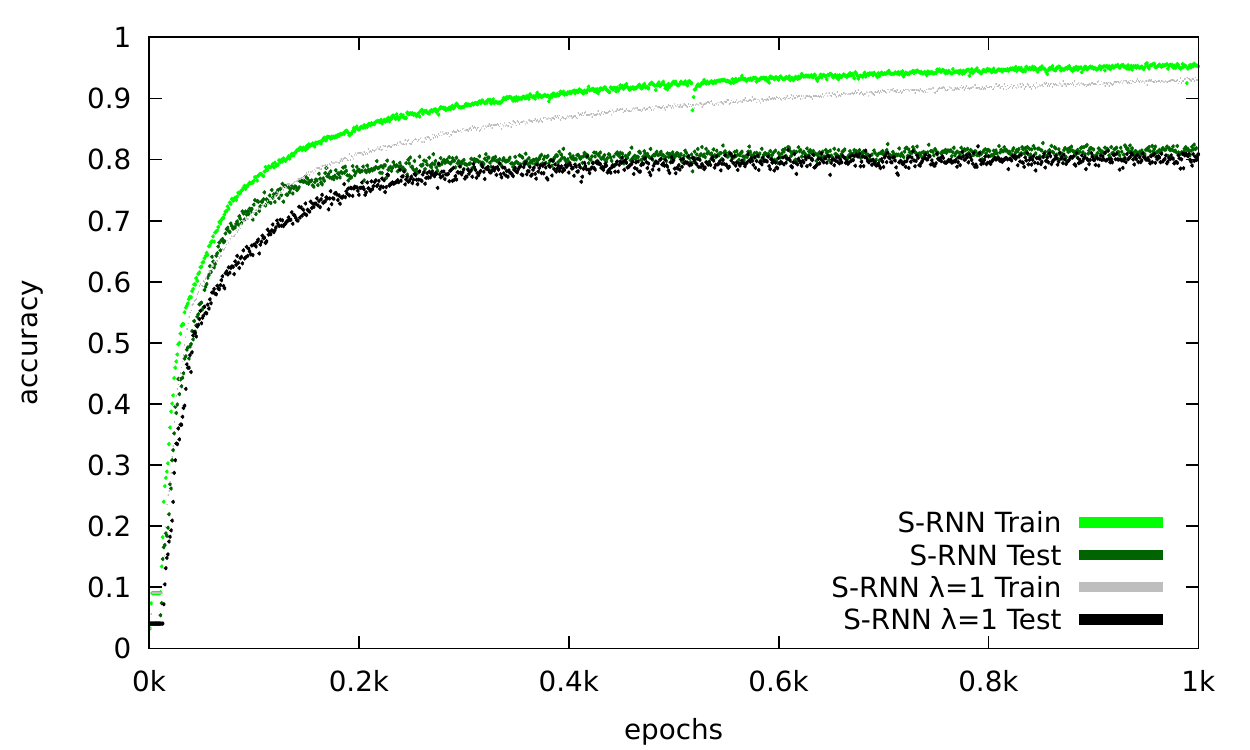}
	        \centering ModelNet40 
	\end{minipage}
	\caption{
		Train and test set learning curves on the point cloud experiments with 500 training samples.
	}
        \label{fig:pointcloudlearning_10pointcloudlearning_40}
\end{figure}

The central result of our paper of generalization on the structure can be seen in figure \ref{fig:pointcloudlearning_10pointcloudlearning_40} where we 
see that even if in training and testing there is an infinite amount of instances, our model is able to express and learn the correct concept.
We also see that using or not the regularizer does not change the result.
\note[Alex-NIPS19]{Not sure I get what is the point you want to make here. I use the same figure to make a discussion on 
the overfitting avoidance in the main text.}
\note[Alex-NIPS19]{What is the performance measure you trace in the curves? the y axis says correlation. I thought it was accuracy on the test set.
Describe it} 

In this experiment, we can dissociate the problem of understanding what the structure of set is from the problem of  
classifying. Our model generalizes almost perfectly on random samples of a given set as it can be seen by its training 
results on instances that it has never seen. However, it does not capture well what associates a 3d model with its respective 
label.  This association can be described by notions such as translation invariance and local feature extractors as the ones
learned by CNNs, which are in fact the methods that achieved the best performance in the classification of 3d cloud points.

%

\subsection{Tree problems}
\label{sec:tree}

%

\subsubsection{Data structures in tree datasets}
We experiment with \strucLearn on two different learning tasks involving instances that are represented as trees. 
We consider ordered and unordered trees.
In the former the order of the child nodes is important while in the 
latter it is not. Ordered trees are mostly used in NLP tasks while unordered are used in graph modeling. 
The first task is a conditional tree generation task in which the goal is to generate a tree given itS textual description.
Thus here a learning instance is a $\mathbf x, \mathbf Y$, pair  where the predictive component $\mathbf x$ 
is a sequence and the target $\mathbf Y$ is a tree. The second task is a regression task where the goal is to 
predict a scalar given a tree, i.e. learning instances are now of the form $\mathbf X,y$, where $\mathbf X$ is a tree
and $y\ \in \mathbb R$. As is customary in (generative) tree modeling we make the assumption that the probability
distribution that governs the generation of a given node is a function of its parent nodes and its so-far seen siblings.
For the conditional tree generation task we use the synthetic data and the evaluation code of \cite{alvarez-melis_tree-structured_2016}.
The goal is to predict the topology of an ordered tree given only its nodes sequence as this is produced by a depth first 
traversal of the tree and no topological information. 
Node labels are taken from the 26-letter alphabet ${\sf T}_1 = \{{\sf A}, 
{\sf B}, \dots {\sf X}, {\sf Y}, {\sf Z}\}$. We use the train/validation/test
set separation of the original paper, i.e. 4000 training, 500 validation, and 500 testing instances. The tree sizes vary considerably
with the smallest trees having only a single node and the largest ones 20, with the average number of nodes  being 4. 
For the regression task the goal is to predict the boiling point of alkane molecules (\cite{DBLP:journals/ijon/GallicchioM13}). Here we consider both ordered and unordered trees. The node labels are taken 
from the set of ${\sf T}_2 = \{{\sf C},{\sf CH},{\sf CH2},{\sf CH3},{\sf CH3F},{\sf CH4}\}$ where each label indicates how many 
hydrogen atoms are linked to the carbon atom. Here the number of nodes per tree vary from one to ten, with an average of five. 
The dataset has 150 learning instances. We estimate the performance by averaging the performance estimates
over three hold-out sets, where the size of the hold out is 20. From the remaining 130 instances we use 100 for training 
and the remaining 30 for parameter tuning. 
\note[Alexandros-NIPS]{Not clear, each node in the tree is what? an atom? and if yes what kind of atom?
only carbons? And what is CH3F?} 
\note[Alexandros-NIPS]{Also, why you consider both ordered and unordered? and what about the first task did you 
consider there too both types of trees?}


\subsubsection{Serialisation for tree problems} 
\label{sec:tree-serialisation}
We now describe how the \strucAlgo treats the learning instances, i.e. the ($\mathbf x, \mathbf Y$) or ($\mathbf X, y$) pairs, starting by 
describing the serialisation of a tree. In serialising tree structures the dictionary $\mathbb B$ contains only categorical elements, and in 
particular it is the set $\{{\sf (}, {\sf )}\} \cup \text{NL} $, where a ${\sf (}$ indicates that the next element of the serialisation 
will be the children of the current node, ${\sf )}$ indicates that we have completed the list of children of the current node, $\text{NL}$ is 
the set of all node labels for the given tree problem, i.e. ${\sf T}_1$ for the tree prediction problem and ${\sf T}_2$ for the boiling point 
prediction problem. To produce the serialisation we traverse the tree in a depth first manner and add elements to the serialisation 
as we move from node to node. For ordered trees the order of traversal of the children of a node is the same as the one 
given by the tree, i.e. there is no randomness here. For unordered trees the order of traversal is random, i.e. $\mu$ 
is uniform over the non-selected children. 
To give an example, for the ordered tree with root ${\sf A}$ and two child nodes ${\sf B}$ and ${\sf C}$   its unique 
serialisation will be $[{\sf A}, {\sf (}, {\sf B}, {\sf C}, {\sf )}]$. If the tree is unordered then it will have two
possible serialisations $[{\sf A}, {\sf (}, {\sf B}, {\sf C}, {\sf )}]$ and $[{\sf A}, {\sf (}, {\sf C}, {\sf B}, {\sf )}]$.
The state $s$ associated with the given partial serialisation we generate just before arriving at some node, $k$,  of a tree will be 
given by the sequence of the parent nodes of $k$ and its so far-seen siblings, i.e. it does not depend on the children of 
its seen siblings. This state representation reflects the main assumption in tree modeling, mentioned above, i.e. that the 
generative distribution of a node is a function of only its parent nodes and its so-far seen siblings. We 
only use this state representations when we want to impose the structural constraints regulariser. The tree 
serialisation is one component of the learning instance serialisation. In the tree prediction problem we need
to serialise $(\mathbf x, \mathbf Y)$ pairs, where $\mathbf x$ is the node label sequence of the depth-first
tree traversal. Since here $\mathbf x$ is already a sequence there is no serialisation involved for it. In addition
since the elements come from ${\sf T}_1$ we do not even need to extend the dictionary $\mathbb B$ since the node 
labels will be already in. However we prefer to use a different label set, $ {\sf T}_1'$,  for the elements of the 
input sequences in order not to provide to the algorithm the domain knowledge about the correspondence of the building blocks
of the sequences and the trees. This makes the problem more difficult since the algorithm will now
need to learn these correspondences. Thus the final dictionary is $\mathbb B = \{{\sf (}, {\sf )}\} \cup {\sf T}_1 \cup {\sf T}_1'$.
The sampling measure $\mu$ we use to serialise a $(\mathbf x, \mathbf Y)$ learning instance randomly selects to 
include first in the serialisation the $\mathbf x$ component half of the times while the other half it first
serialises the tree $\mathbf Y$. Essentially we are feeding the model with samples from both $P(\mathbf Y|\mathbf x)$
and $P(\mathbf x | \mathbf Y)$ distributions and the learning algorithm learns associations between their individual 
building blocks, learning eventually the complete joint distribution $P(\mathbf x, \mathbf Y)$. For the regression
task where the learning instances come in the form $\mathbf X, y$ the dictionary is now given by 
$\mathbb B = \{{\sf (}, {\sf )}\} \cup {\sf T}_1 \cup {\sf t } \cup \mathbb R $ . It thus includes
also real value elements, since these are used for the target variable. The label ${\sf t}$ stands for target
and it will always be followed by a scalar, describing thus the target value $y$ for the given training instance.
As in the tree prediction problem the sampling measure $\mu$ selects randomly in half the serialisations the $\mathbf X$ 
component first and in the other half the $y$ component. 

\note[Alexandros-NIPS]{I wonder now whether the fact that you use DFS does explain the very good performance on 
the first tree dataset.  Since your tree representation is almost equivalent to the task input.}

\subsubsection{Learning architecture for tree problems} 
On the conditional tree generation task we compare our method against DRNN introduced 
in \cite{alvarez-melis_tree-structured_2016} using the authors' code
and their evaluation protocol. DRNN use two different hidden state vectors a fraternal and an ancestral.
The fraternal hidden-state models the evolution of the state with siblings and the ancestral models 
the relation between  parent and child. This relation is modeled with two types recurrence: one between 
parent and child, and one between siblings. The hidden state is then used to predict the topological 
information (if we grow a new branch) and the label information.
The evaluation protocol treats the task as a retrieval problem quantifying the quality of the 
recovery of the nodes and edges of the original tree. We use the same learning architecture as the one described in \ref{sec:dynamical-system}
with small differences in the number of hidden units and layers. Concretely we use a two-layer LSTM with 512 units followed by a two-hidden 
layer network that predicts the categorical component and another two hidden layer that predicts the parameters (means, variances and 
mixture coefficients) of 6 Gaussian mixtures. In the two latter networks the number of hidden units is tuned on the validation  set
from  $2^i:i=5\ldots10$. The $\lambda$ parameter of the structural constraints regulariser is also tuned from the set $(0,1,10,100)$. We report choose the best model on the validation set and report the testing error.
As before we use ADAM for optimization. We use a mini-batch size of 32 instances for DRNN. 
For \strucLearn a mini-batch contains 64 serialisations which are generated from 32 instances.

\subsubsection{Tree results}
We give the results in table \ref{TreeSynth} for the conditional tree generation task. 
\strucLearn outperforms DRNN by a large margin, a method specifically developed to learn with trees, 
both for the F1 and precision measures for nodes and edges. It fairs worse for node and edge recall. DRNN has a perfect recall, at the cost of 
generating trees which have many superfluous elements.

\begin{table}[htbp]
\begin{tabular}{ccccccc}
\hline
Algorithm&Node F1&Node precision&Node Recall&Edge F1& Edge Precision&Edge Recall\\ \hline
\strucLearn         & 88.95\% &87.82\%&90.79\%&83.43\%&82.22\%&85.47\%\\
DRNN                & 74.51\% &59.37\%&100\%&65.86\%&49.10\%&100\%\\ \hline\\
\end{tabular}
\caption{Performance results for the conditional tree generation task.} \label{TreeSynth}
\end{table}

In the regression task we compare \strucLearn against two variants Tree Echo State 
Network,  TreeESN-R and TreeESN-M, \cite{DBLP:journals/ijon/GallicchioM13}. Both TreeESN methods are reservoir computing 
models which generalize the reservoir computing paradigm to tree structured data. The difference between the two variants
is on how they aggregate the state vectors to represent the complete tree. The R variant only uses the state of the root 
whereas the M variant averages over all states of the tree.  We experiment with ordered 
and unordered trees. The evaluation error is the mean absolute error. We give the performance results in 
table~\ref{TreeReg} (average predictive error).  The two variants of \strucLearn, i.e. trained on ordered and unordered trees, give 
better results than TreeESN-R , while they perform worse than TreeESN-M. The performance 
of all methods is quite remarkable given that the scalar values to predict range from  \SI{-164}{\celsius} to \SI{174}{\celsius}.

\begin{figure}
	\includegraphics[width=0.3\linewidth]{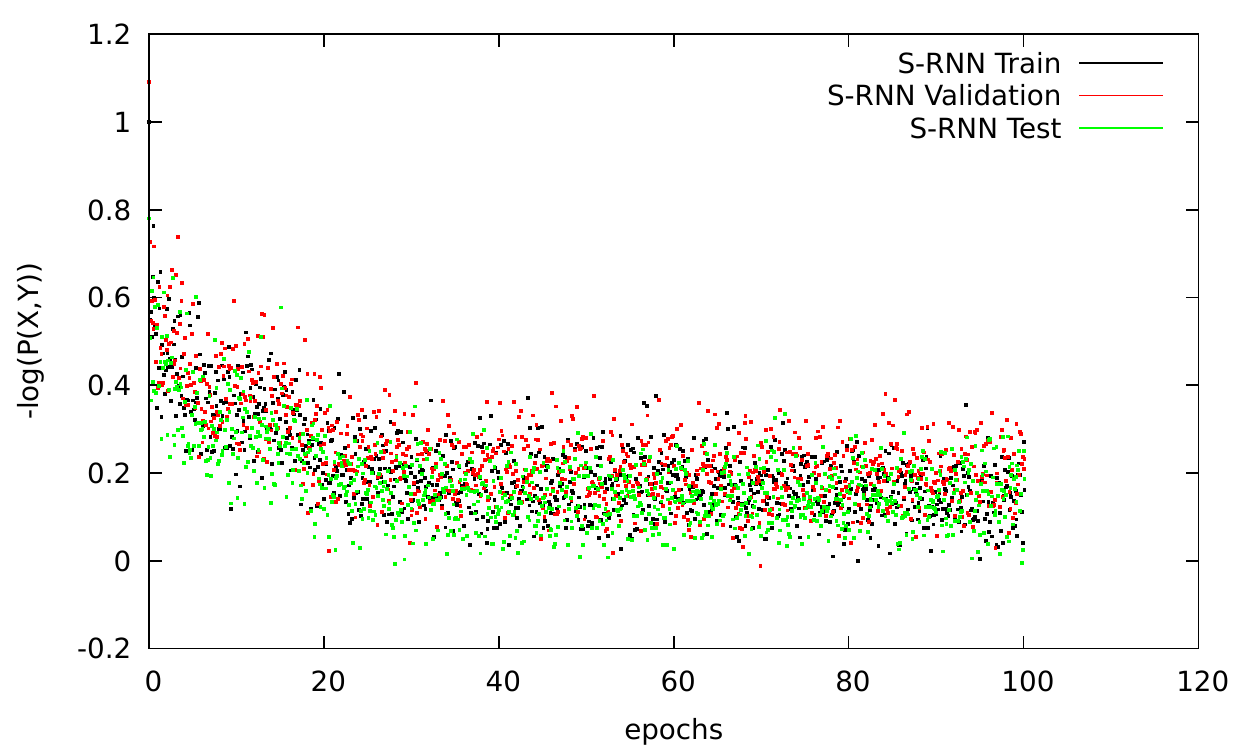}
	\includegraphics[width=0.3\linewidth]{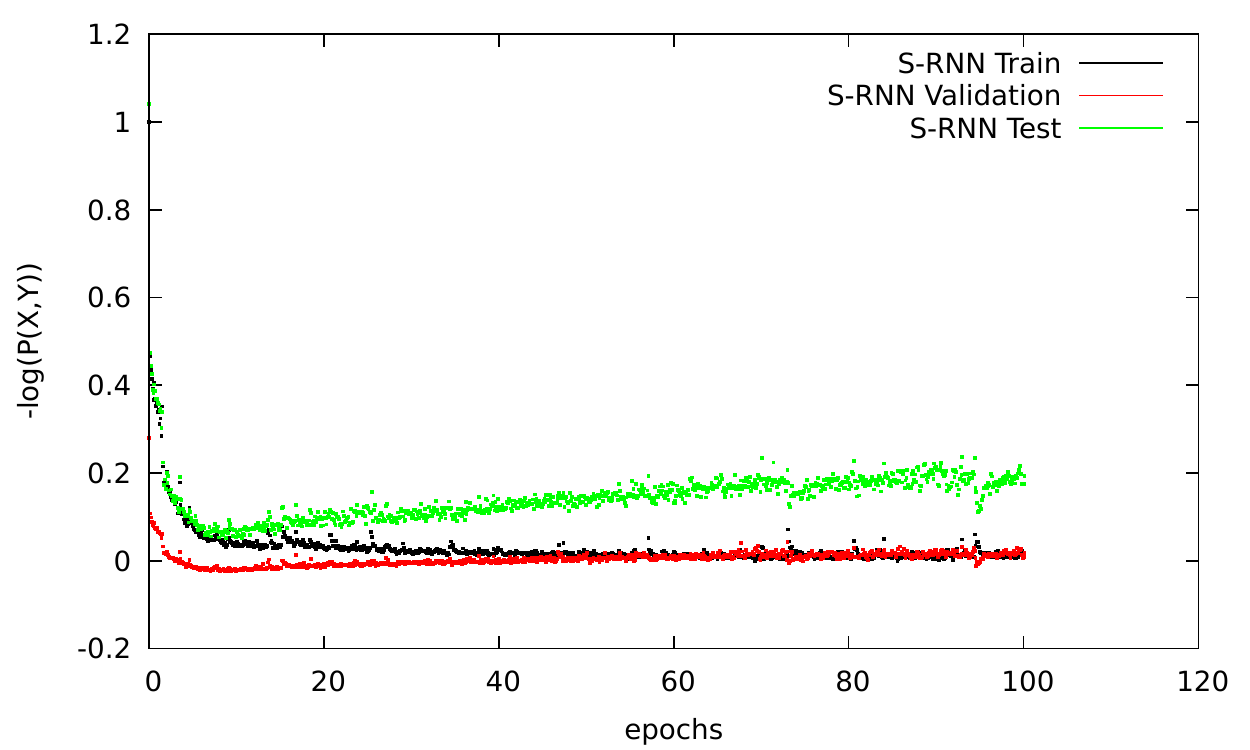}
	\includegraphics[width=0.3\linewidth]{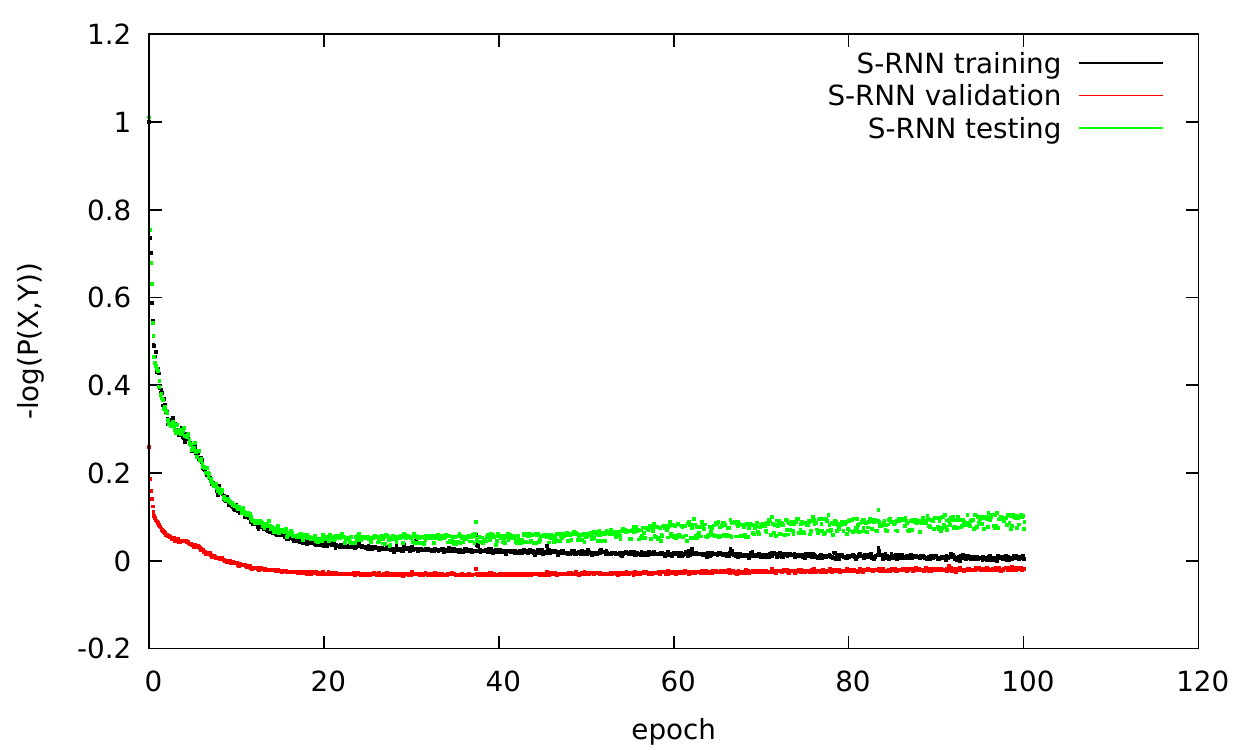} 
	\caption{Learning curves in training/validation/test sets as a function of the nunber of epochs, tree problems. Conditional 
tree generation (left), tree regression, ordered trees, (middle), tree regression, unordered trees, (right).}
	\label{fig:trees-lossevolution}
\end{figure}

In order to check the behavior of \strucLearn with respect to overfitting we also plot the evolution of the loss 
in the train/validation/test sets in figure \ref{fig:trees-lossevolution} as a function of the training epoch number. 
When it comes to the conditional tree generation task and the tree regression task with the unordered trees, there is hardly 
any divergence between the training, validation, and test losses. In the case of ordered trees we do observe an important 
divergence starting from around the 20th epoch. 
In serialising an ordered tree there is no randomness since we have to respect the order, thus an ordered tree has a single 
serialisation, contrary to the unordered ones which have multiple serialisations. Exposing the learning algorithm to multiple 
random, but equivalent, serialisations provides clear benefits in terms of protection against overfitting. 

\begin{table}
	\begin{center}
	\begin{tabular}{lc} \hline
		Algorithm & Error\\ \hline
		\strucLearn ordered trees   &\SI{7.18}{\celsius}\\
		\strucLearn unordered trees &\SI{6.15}{\celsius}\\
		TreeESN  M & \SI{2.78}{\celsius}\\
		TreeESN  R & \SI{8.09}{\celsius}\\ \hline
	\end{tabular}
\caption{Average predictive error on the tree regression task.}
\label{TreeReg}
	\end{center}
\end{table}

\section{Graph/Molecule problems}
\label{sec:graph}
We experiment with \strucLearn on a set of regression tasks and datasets where instances are graphs. 
We use the QM9 dataset from deepchem~\citep{Ramsundar-et-al-2019} benchmark and the Guacamole dataset 
from \cite{brown_guacamol:_2018}. In both cases the goal is to predict a number of properties from 
the molecule structure. For the QM9 dataset, the regression problems/targets, are given in the dataset
and are:
mu, alpha, HOMO, LUMO, gap, R2, ZPVE, Cv, U0 u298, h298, g298. 
For the Guacamole dataset, we computed the following targets using RDKit: logP, mol\_weight, num\_atoms, num\_H\_donors, tpsa.
\note[Alex-NIPS19]{Who did the computation for Guacamole? you? if yes say it explicitly. Also it might be
good to list the targets for the two datasets.}
In the publicly available versions of the two datasets the molecules are represented
by their canonical SMILES strings \citep{weininger_smiles_2002}. Our randomised serialisations are 
generated by exporting the SMILES strings using RDkit \citep{rdkit} with the randomize option on.  
The vocabulary of our alphabet are the symbols of the SMILES strings and we use a one hot encoding.
We measure error with the squared Pearson correlation coefficient. 
As before we learn over the sequences using a GRU\citep{cho_learning_2014} with 128 units, 
and use over it a single layer with 128 hidden units to solve the regression task. We did
not use regularisation ($\lambda=0$). As optimizer we use Adam\citep{DBLP:journals/corr/KingmaB14} 
with a learning rate of $10^{-4}$.

To compute the baseline results on the QM9 dataset 
we used the benchmark script from deepchem~\citep{Ramsundar-et-al-2019} with the default options. We give 
the complete results in tables \ref{result:QM9-Full-1} and \ref{result:QM9-Full-2}.
Below is a short description of the baselines we used for the QM9 task: \note[Alex-NIPS19]{Provide references for all your baselines. 
Did you implement them? took results from papers? be explicit}
\begin{description}
	\item[tf regression] MultitaskDNN is a standard MLP that predicts multiple tasks. \note[Alex-NIPS19]{I think you can say something more about this and the next one.}
	\item[tf regression ft] Fit Transformer MultitaskDNN is a variant of the previous which in addition does a binarization 
and transformation of the input feature. We used the best domain knowledge guided transformation on this dataset as determined by the benchmark author\citep{Ramsundar-et-al-2019}.
\note[Alex-NIPS19]{What does it mean that they are cross validated?}
	\item[graph conv reg] Graph convolution regression is a variant of graph convolution which produces a 
fingerprint of the molecule that is then used by a classifier.
	\item[weave regression] Weave is a variant of graph convolution which does the convolution on the whole molecule.
	\item[dtnn] Deep Tensor Neural Network uses as an additional feature the 3D coordinates of the atoms. 
Using these coordinate an update mechanism based on the distance matrix and the neighborhood is used. This is 
the only model that uses the 3d coordinate.
\end{description}

For the Guacamole dataset, we did not found an existing result in a regression settings. All 
baselines we found were in the setting of generation. We decided to still included the results 
of Guacamole to show the scalability of our method. We give the complete results in table \ref{result:Guacamole-app}.

An interesting particularity of the mu task in QM9 is that all models that use only the graph and 
no 3D position information perform signigicantly worse compared to dtnn which uses the 3D position information.
For the Guacamole experiment we see that we almost perfectly predict the target. This is not too surprising as 
the target was computed with RDkit and is quite simple.  From these experiments we see that even with a simple 
sequence model we can have state of the art performance on learning tasks as complex as molecule properties prediction.

\begin{figure}[hp]
	\includegraphics[width=0.45\textwidth]{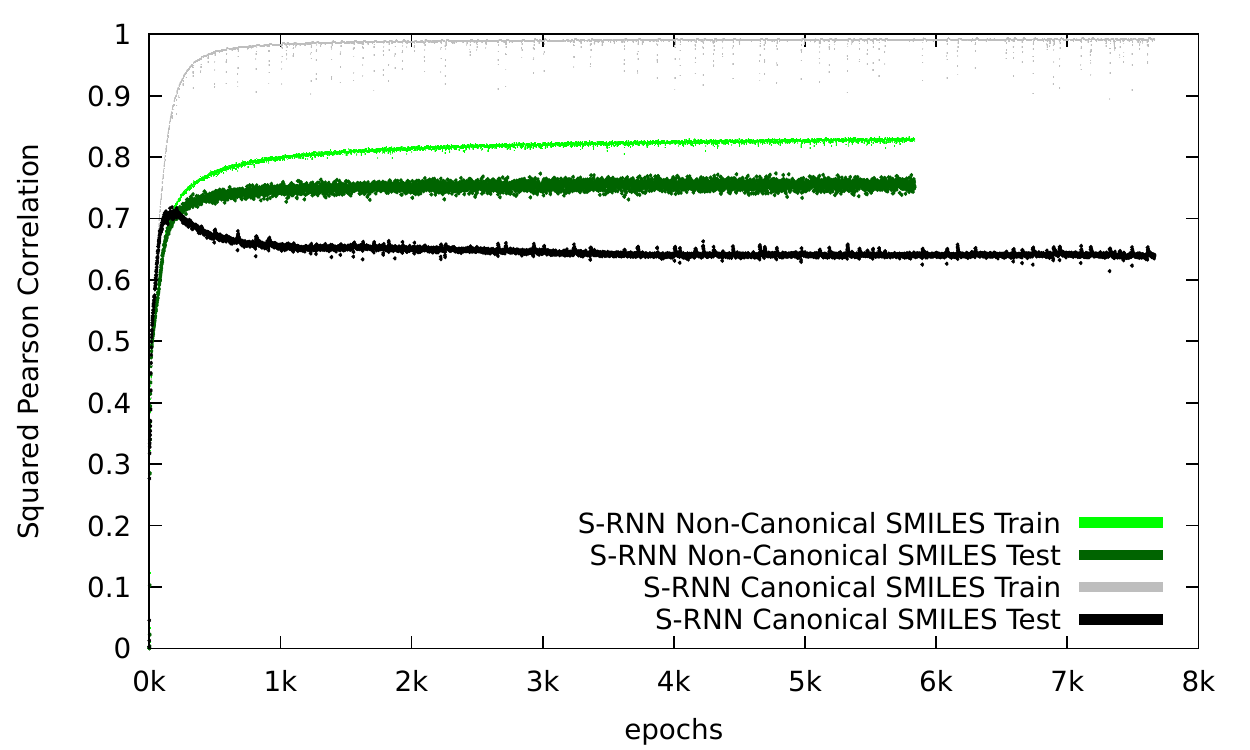} \includegraphics[width=0.45\textwidth]{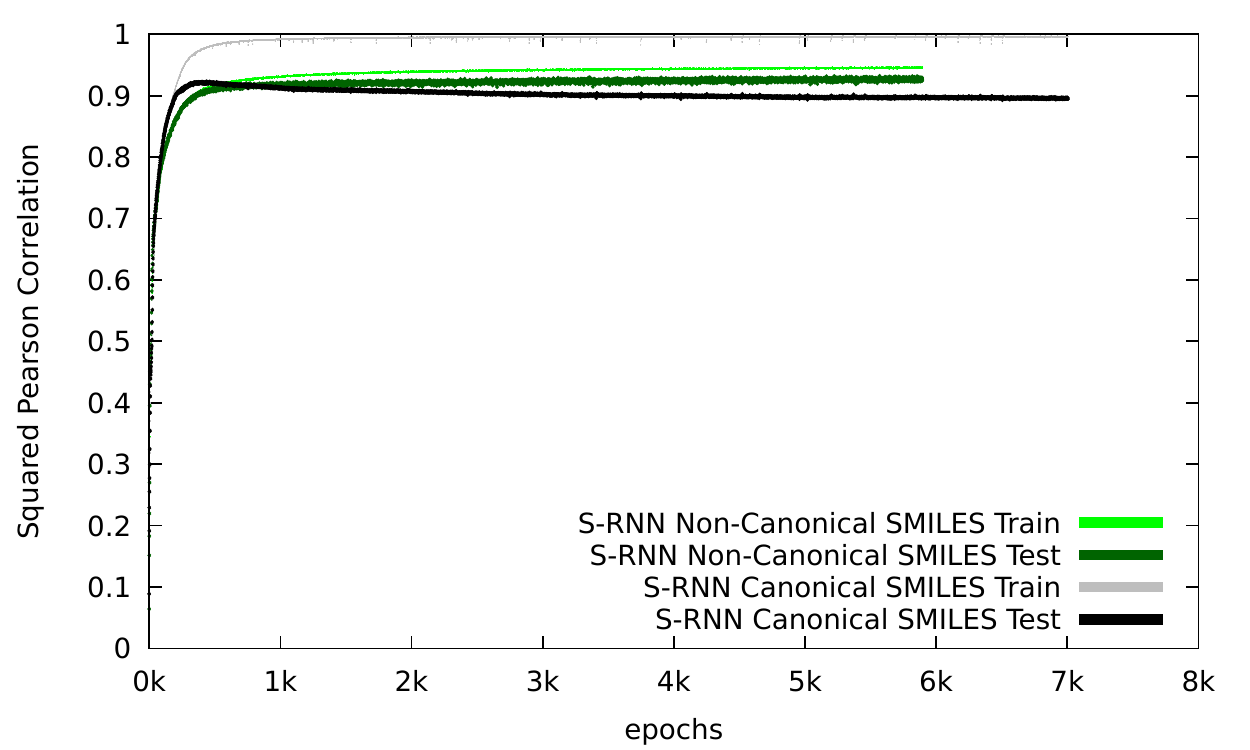}
	\caption{Learning curves of the QM9 dataset for two different tasks. Left: mu task. Right: homo task.}
	\label{fig:MoleculeMuLearning}
\end{figure}

The learning curves on the mu task of QM9 on the canonical and non-canonical SMILES provide us with an eloquent 
demonstration of the benefits of the randomization (figure \ref{fig:MoleculeMuLearning}). There we see that in the randomized SMILES the learning
curves on the training and test exhibit very similar relative behaviors even after 6k epochs. 
This is not the case for the canonical smiles where very early the train and test learning 
curves diverge.  Without randomization, the model may overfit on a particular 
SMILES string. With randomization the additional complexity of all possible equivalent 
SMILES strings forces the model to generate a representation which is compatible with 
the randomization procedure.

\begin{table}
\begin{center}
	\small
	\begin{tabular}{l|c|c|c|c|c|c|c|c|c|c|c|c}
		\hline
		Algorithm                            &mu  &alpha&HOMO &LUMO  &gap  &R2       \\
		\hline
		\strucLearn Non-Canonical SMILES     &0.75&0.993& 0.93& 0.985& 0.97& 0.970   \\
		\strucLearn Canonical SMILES         &0.64&0.994& 0.90& 0.980& 0.95& 0.968   \\
		tf regression                        &0.72&0.659& 0.86& 0.949& 0.93& 0.734   \\
		tf regression ft                     &0.73&0.963& 0.84& 0.939& 0.92& 0.977   \\
		graph conv reg                       &0.76&0.810& 0.91& 0.977& 0.96& 0.824   \\
		weave regression                     &0.71&0.954& 0.89& 0.966& 0.94& 0.947   \\
		dtnn                                 &0.97&0.993& 0.96& 0.988& 0.98& 0.998   \\
		\hline
	\end{tabular}
	\caption{Predictive performance on the QM9 dataset. Correlation between predicted and real value.}
	\label{result:QM9-Full-1}
\end{center}
\end{table}

\begin{table}
\begin{center}
	\small
	\begin{tabular}{l|c|c|c|c|c|c|c|c|c|c|c|c}
		\hline
		Algorithm                        &ZPVE    &Cv  &U0     &u298   &h298u   &g298\\
		\hline
		\strucLearn Non-Canonical SMILES & 1.000& 0.995& 1.000 & 1.000 & 1.000& 1.000\\
		\strucLearn Canonical SMILES                                              & 1.000& 0.994& 1.000 & 1.000 & 1.000& 1.000\\
		tf regression                                                             & 0.880& 0.741& 0.671 & 0.671 & 0.671& 0.671\\
		tf regression ft                                                          & 0.988& 0.978& 0.994 & 0.994 & 0.994& 0.994\\
		graph conv reg                                                            & 0.927& 0.824& 0.741 & 0.741 & 0.741& 0.741\\
		weave regression                                                          & 0.982& 0.963& 0.984 & 0.984 & 0.985& 0.985\\
		dtnn                                                                      & 0.999& 0.997& 0.998 & 0.998 & 0.998& 0.998\\
		\hline
	\end{tabular}
	\caption{Predictive performance on the QM9 dataset. Correlation between predicted and real value.}
	\label{result:QM9-Full-2}
\end{center}
\end{table}

\begin{table}
	\small
	\centering
	\begin{tabular}{l|c|c|c|c|c}
		\hline
		Algorithm                 &logP&mol\_weight&num\_atoms&num\_H\_donors&tpsa\\
		\hline
		\strucLearn Non-Canonical SMILES        &0.999&0.999  &0.999  &0.999&0.999\\
		\strucLearn Canonical SMILES            &1.000&0.999  &0.999  &0.999&0.999\\
		\hline
	\end{tabular}
\caption{Predictive performance on the Guacamole dataset, Pearson correlation coefficient with the target.}
\label{result:Guacamole-app}
\end{table}

\note[Alex-NIPS19]{Where is the table with the results for Guacamole, and why there are no baselines there? explain}

\label{Graph-Long}

\note[Alex-NIPS19]{Change labels from "Train Randomized" -> "Train with Randomized SMILES", and so on. Place them next to each other.}

\subsection{Multivariate dynamical systems datasets}
\label{sec:dynamical-system}

\subsubsection{Data structure of multivariate dynamical systems} 
We explore the performance of \strucLearn on two datasets arising from multivariate dynamical systems (artificial and real-world).
We generate the artificial dataset using a known dynamical system.
The real world dataset contains recordings of gait trajectories (joint angle values) of people with pathological gait. Both datasets have a
similar structure and differ only by their dimensionalities and cardinalities.  In particular every 
training instance consists of two components: $\mathbf x \in \mathbb R^d$, which we call input, and 
$\mathbf Y \in \mathbb  R^{k \times l}$ which we call output; with the latter being a probabilistic 
function of the former. We will denote the $i,j$ element of $\mathbf Y$ by $y_{ij}$, the $j$ column by $y_{.j}$
and the $i$ row by $y_{i.}$; functions $l(y_{i.})$ and $l(x_i)$ return the name of the feature they take as argument.
\note[Alexandros]{ok... so in the data we have a dynamical system part that is of fixed dimension. Not nice.} The
$\mathbf Y$ matrix contains a $k$-dimensional dynamical system uniformly sampled at $T$ time points. We
solve two types of tasks. A conditional generation task in which the goal is to learn the conditional density 
$\Pr(\mathbf Y| \mathbf x)$ and use that for sampling and prediction  and an unconditional generation task in which 
we seek to learn $\Pr(\mathbf Y)$ and sample from it. In both cases we measure performance with the negative log-likelihood.

\subsubsection{Serialisation of multivariate dynamical systems} 
We now describe the concrete serialization structure that the \strucAlgo produces for a particular 
$\mathbf Y$ matrix and a $\mathbf x, \mathbf Y$, couple. Our dictionary $\mathbb B$ contains two types of elements, 
categorical and real valued. The domain of the categorical elements is 
$\{l(x_i)|i := 1...d\} \cup \{l(y_{i.})|i := 1...k\} \cup \{t+\}$, i.e. the
names of the features of the $\mathbf x$ and $\mathbf Y$ components and $\text{t}+$; the latter 
denotes a shift from a column of the $\mathbf Y$ matrix to the next one, essentially 
it corresponds to moving to the next element of a multi-variate sequence. The real valued elements are the values of the features. 
Within a serialisation a categorical element is always coupled by a real value. A
feature name is coupled by the respective feature value and $t+$ is always coupled with zero. The categorical elements are encoded with a one-hot vector. 

When serializing matrix $\mathbf Y$ and currently at column $j$ the \strucAlgo
 randomly chooses among the features that have not yet been added which one to 
 add.
 Thus $\mu$ is uniform over the non-selected features. Once all features of the $j$ column 
have been sampled then the $t+$ operator is selected as the next element of the serialisation, 
and the \strucAlgo proceeds with the serialisation of the next sequence element. When we 
serialise an ($\mathbf x, \mathbf Y$) couple the sampling measure $\mu$ 
is now different. In half of the cases we first select all elements of the $\mathbf x$
component to be added to the serialisation before moving to the serialisation of the $\mathbf Y$ component.  In 
the other half sampling between the $\mathbf x$ and $\mathbf Y$ components is uniform, i.e. $\mathbf x$ 
and $\mathbf Y$ features can be interleaved.
Nevertheless the serialisation order of $\mathbf Y$ is the same 
as before. We bias serialisation towards selecting first the $\mathbf x$ components because we want to sample and learn 
the conditional distribution $\Pr(\mathbf Y| \mathbf x)$ thus the conditioning component should appear
first in the serialisation. However, we still allow for a uniform sampling between the $\mathbf x$ and $\mathbf Y$ components
in half of the cases so that the learner will have more chance to pick up on correlations between parts 
of the input and parts of the output. All serialisations are generated on the fly during training. 
\note[Alexandros]{And why not all the time?}
\note[Pablo]{To allows the model to learn correlation between input and output.} 
\note[Alexandros]{I added the answer but to me it make no sense.}

\subsubsection{Learning architecture for multivariate dynamical systems} 
We describe the learning architectures we use. Note that these architectures are essentially 
the same for the baseline learning algorithms (against which we will compare) and our algorithm \strucLearn.  
The architectural differences are only the result of the structure of the training data. In the case of the baseline algorithms these
are either standard vectorial data, i.e. here the $\mathbf x$ component, or a $k$-dimensional sequence, i.e.
the $\mathbf Y$ component. In the case of \strucLearn the training data are the serialisations/sequences 
produced from a given training instance $\mathbf Y$ or ($\mathbf x, \mathbf Y$), where each serialisation element is
a couple with a categorical component and its respective real value. 
\note[removed]{The differences between the 
baseline and the \strucLearn architectures come from the adaptation of the dimensionalities of the input units to accommodate
for the different structures/dimensions of the training data and the fact that serialisation elements have a categorical 
component which must also be predicted. }

We first describe the baseline architecture. We model the probability of the next $k$-dimensional element in a sequence 
given the current state as a $m$-component mixture of Gaussians the parameters of which we learn. Both for the unsupervised and supervised 
case the core architectural element is a multivariate LSTM.  For the unsupervised setting we use a two-layer 
LSTM (\cite{LSTM}), with 128/256 units in each layer for the artificial/gait datasets respectively, followed by 
a one hidden layer neural network with 128/64 units for the artificial/gait datasets respectively.
The network is fed sequentially with the $k$-dimensional sequence of the $\mathbf Y$ matrix
and predicts the  means, covariance matrices, and mixture weights of the Gaussian mixture
 (thus its output is of dimensionality $ m\times(k + k \times k) + m$), which provides the conditional 
 distribution of the next sequence element. 
For the artificial data the mixture has only 1 Gaussian component and for the gait data it has 6. 
For the supervised setting we use an encoder-decoder architecture built on top of the architecture we just 
described (\cite{NIPS2014_5346}). The encoder part has the same architecture as the two-layer LSTM we just described
and is fed with the $\mathbf x$ component, i.e. a single element sequence. The hidden states and cell states
of the two layer encoder are fed to the respective states of the decoder which itself also has the same two layer 
architecture and as in the unsupervised case feeds to a single layer neural network.
All dimensionalities are the 
same as before. 

For \strucLearn since each element of the serialisation has a categorical component and a continuous one we need to 
adapt the learning architecture for that structure. We use exactly the same architecture for the supervised and unsupervised
experiments since there is no change in the serialisation structure between the two experiments. 
To adapt the baseline architecture we described in the previous paragraph to the particularities of the 
serialisation structure we add one more one hidden 
layer network which is fed by the output of the two-layer LSTM and together with a soft-max layer model the conditional
probability of the categorical part of the next element in the serialisation. The continuous component is predicted 
using the same architecture as the one we describe before to predict the $k$-dimensional element of a sequence, with 
the only difference that since it is a scalar the output of the network will have $m \times (1 + 1) + m$ outputs predicting
the mean, variance and mixture weights of the $m$ component Gaussian mixture. 
\note[Alexandros]{Nice one. }

We optimize all architectures using Adam (\cite{DBLP:journals/corr/KingmaB14}). We use a mini-batch size of 32 instances for 
the baseline methods. In the case of \strucLearn a mini-batch contains 64 serialisations which are generated from 32 instances.

\note[Alexandros]{@Alexandros Regulariser, encoding of operators.}

\subsection{Artificial dynamical system}
We use a couple of Van der Pol equations linked to an harmonic oscillator to generate the artificial 
dynamical system. The coupling creates correlations between the variables which the learning process
needs to learn. Here the dimensionality $d$ of $\mathbf x$ is 9, and the dimensionality of $\mathbf Y$ is 
$3 \times 21$. Given an input $\mathbf x$, its matrix $\mathbf Y$ is generated by:
\begin{align*}
\frac{d^2y_1(t)}{dt^2}&=-|k|_{+} y_1(t)\\
\frac{d^2y_2(t)}{dt^2}&=|\mu_y|_{+}(1-y_2(t)^2)\frac{dy_2(t)}{dt}-y_2(t)-y_1(t)\\
\frac{d^2y_3(t)}{dt^2}&=|\mu_{y_3}|_{+}(1-y_3(t)^2)\frac{dy_3(t)}{dt}-y_3(t)-y_2(t)
\end{align*}
The input vector $\mathbf x$ contains the initial conditions of the dynamical system and the values of its parameters, 
$y_1(0)$,$y_2(0)$,$y_3(0)$,$\dot{y_1}(0)$,$\dot{y_2}(0)$,$\dot{y_3}(0)$,
$k-3$,$\mu_{y_2}-3$ and $\mu_{y_3}-3$. These are generated randomly for each ($\mathbf x, \mathbf Y$) pair. 
We generated 3000 instances of length 21 which we divided equally to training, validation, and testing sets. 
We train for 12 hours or until the validation error becomes larger than the validation error of the first iteration, which in 
the case of the baseline happens very often. We then select the model with the lowest validation error and apply it on 
the test set to compute the conditional negative log-likelihood. With the artificial 
dynamical system we only experiment in the conditional generation setting; the generated $\mathbf Y$ 
component is the one that maximizes the $\Pr(\mathbf Y|\mathbf x)$ conditional likelihood. 

\begin{figure}[htbp]
	\centering
	\begin{minipage}{0.32\linewidth}
		\includegraphics[width=\linewidth]{Likelihood.pdf}
		\centering Artificial dynamical system, supervised setting.
	\end{minipage}
	\begin{minipage}{0.32\linewidth}
		\includegraphics[width=\linewidth]{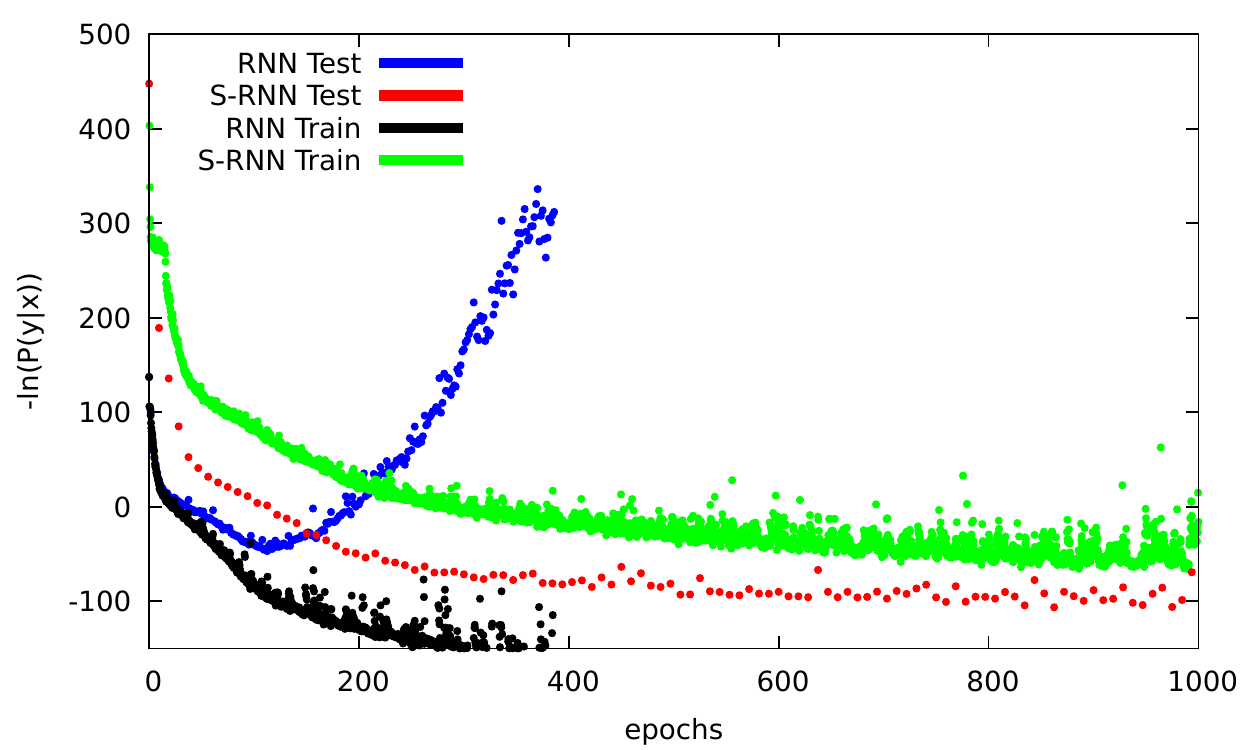}
		\centering Artificial dynamical system, supervised setting.
	\end{minipage}
	\begin{minipage}{0.32\linewidth}
		\includegraphics[width=\linewidth]{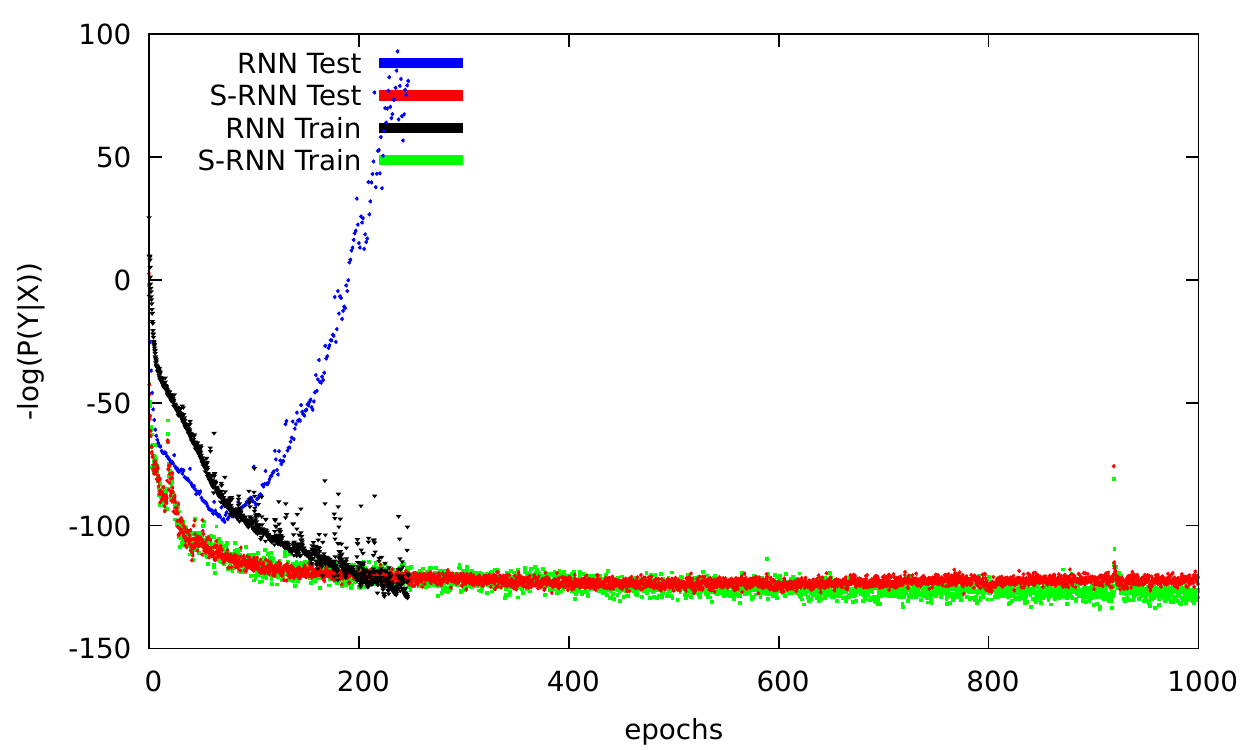} 
		\centering Real world Gait dataset, supervised setting.
	\end{minipage}
	\caption{
	Overfitting behavior on the dynamical system problems, supervised setting. 
	Left: Conditional negative log likelihood on the validation set as a function of training epoch, 
	artificial dynamical system, includes different regularisation levels for \strucLearn.
	Middle: Same as the left but on train/test sets.   
	Right: Same as the middle but on the Gait dataset. 
	}
        \label{fig:simulatedDataSupervisedExpLikelihoodEvolutionAndgaitOverfitting}
\end{figure}

In the left part of 
figure~\ref{fig:simulatedDataSupervisedExpLikelihoodEvolutionAndgaitOverfitting} we give the evolution of the conditional log-likelihood on 
the validation as a function of the number of epoch seen. The most striking observation is that \strucLearn never 
overfits; this is even more clearly demonstrated in the middle graph of the same figure where we give the evolution of 
the likelihood on the train and test set for both \strucLearn and \baseLearn. The standard \baseLearn starts overfitting 
after around 160 epoch. \strucLearn practically will never
see an instance twice due the combinatorial complexity of the serialisation generation and can keep on training practically 
forever and no overfitting. As we can see in Table
~\ref{tab:simulatedDataSupervisedExpLikelihood} \strucLearn with no regularisation
achieves the best result, far better and significantly better than the baseline; we controlled the statistical significance using 
a t-test. Mildly regularising \strucLearn does not seem to bring any performance gain, 
while strong regularisation harms. The fact that regularisation does not bring any effect can be explained by the fact 
that the algorithm never sees twice the same serialisation and thus there is no overfitting problem.

\begin{figure}
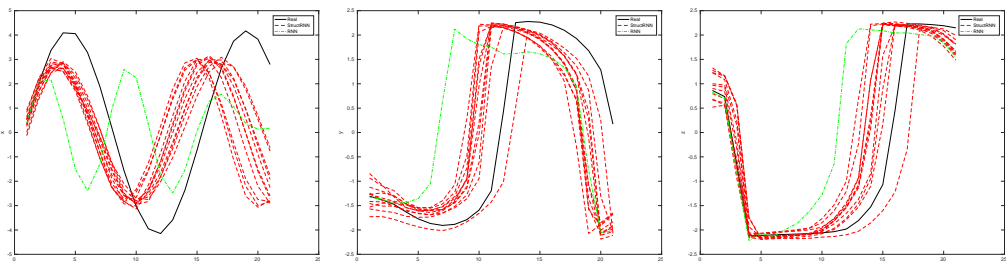

        \begin{minipage}{0.32\columnwidth}
                \includegraphics[width=1\linewidth,trim={2cm 0 3.1cm 0},clip]{Sample_Dyn_System_x.pdf}
        \end{minipage}
        \begin{minipage}{0.32\columnwidth}
        \includegraphics[width=1\linewidth,trim={2cm 0 3.1cm 0},clip]{Sample_Dyn_System_y.pdf}
\end{minipage}
        \begin{minipage}{0.32\columnwidth}
        \includegraphics[width=1\linewidth,trim={2cm 0 3.1cm 0},clip]{Sample_Dyn_System_z.pdf}
\end{minipage}
	\caption{Examples of the $y_1, y_2, y_3$ sequences produced by \strucLearn, \baseLearn for the simulated dynamical system
	 given the $\mathbf x$ component. The black curves are the real data, green is the sequence produced by \baseLearn, and 
	 red are the sequences produced by \strucLearn for different serialisations of $\mathbf x$ input.}
        \label{fig:simulatedDataPredictedY}
\end{figure}

In order to inspect the visual quality of the predictive results we give in figure \ref{fig:simulatedDataPredictedY} 
for a given $\mathbf x$ component the three components of the output sequence which has the maximum conditional 
probability $\Pr(y_1|\mathbf x)$ for \baseLearn and \strucLearn.
As it is obvious \strucLearn produces sequences of better quality, closer to the real sequence. \strucAlgo has 
different predictions as a function of the different serialisations of the $\mathbf x$ component.
\note[Alexandros]{This is tricky since one can ask which one should I pick then...}

\begin{table}
{
	\begin{center}
	\begin{tabular}{|c|c|c|c|c|c|c|c|}
		\hline
		\strucLearn      & \strucLearn $ \lambda=10^2$ & \strucLearn $\lambda=10^4$ & \baseLearn\\ \hline
		$\mathbf{-107}$ & $\mathbf{-103}$            & -57             & -41\\ 
		\hline
	\end{tabular}
\end{center}
}
\caption{Test set negative log likelihood, artificial dynamical system, supervised setting. 
	Bold indicate performances that are significantly better than the non-bold.}
\label{tab:simulatedDataSupervisedExpLikelihood}
\end{table}

\subsection{Gait data}
The gait dataset contains data for 806 patients. Every patient has an $\mathbf x$ component 
which is a 212-dimensional vector describing clinical properties of the patient, related to their body geometry
and articulation flexibility. The $\mathbf Y$ component is an 8-dimensional sequence with 34 observations. The sequence
describes a complete gait cycle of the patient, uniformly sampled at 34 points. Each one of the dimensions is an angular measurement
on a joint of the patient. For each patient we have on average 6 gait cycles, giving a total of 4680 cycles. We decided to define 
learning instances on the level of cycles, thus we have a total of 4608 instances, all of which have an $\mathbf x$ and $\mathbf Y$. 
As a result patients can appear multiple times (depending on how many cycles they have), their $\mathbf x$ component is always the same.
When dividing in training, validation and testing sets, we took care to put all instances of a given patient only in one of the three 
sets. The training set contains 408 patients and their 3276 cycles, the validation set contains 16 patients and 1404 cycles, 
and the testing 382 patients and 1404 cycles. The stopping rule is the same as in the artificial dynamical system. 

\begin{figure}
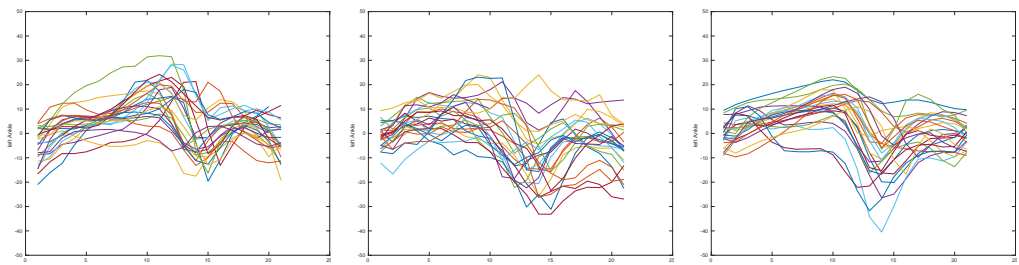

	\begin{minipage}{0.32\columnwidth}
		\includegraphics[width=1\linewidth,trim={2cm 0 3.1cm 0},clip]{Sample_L_Ankle.pdf}
	\end{minipage}
	\begin{minipage}{0.32\columnwidth}
	\includegraphics[width=1\linewidth,trim={2cm 0 3.1cm 0},clip]{Sample_L_Ankle_Rnn.pdf}
\end{minipage}
	\begin{minipage}{0.32\columnwidth}
	\includegraphics[width=1\linewidth,trim={2cm 0 3.1cm 0},clip]{True_L_Ankle.pdf}
\end{minipage}
	\caption{ Left graph: Samples from \strucLearn for the unsupervised Gait problem.
	Middle graph: Samples from \baseLearn for the unsupervised Gait problem.
	Right graph: Real gait samples.}
	\label{fig:GaitSamplesStructRNNAndGaitSamplesRNNAndRealGaits}
\end{figure}

We first report the results on the unconditional generation in which our goal is to learn a model of $\Pr(\mathbf Y)$. 
In table~\ref{tab:gaitDataUnSupervisedExpLikelihood} we give the negative log-likelihood on the test set for different dimensionalities 
of the gait sequence. As it is clear \strucLearn achieves a performance which is always considerably better than the \baseLearn baseline.
\note[Alexandros]{And the natural question here is why not report also on the regularised versions?}
In figure~\ref{fig:GaitSamplesStructRNNAndGaitSamplesRNNAndRealGaits} we give examples of samples generated
from \strucLearn, \baseLearn and real gait cycles respectively. Although the graphs are not conclusive it seems
that \strucLearn preserves more of the real gait cycle structure, while the ones generated from \baseLearn seem to have
a more random structure.

\begin{table}
{
		\begin{center}
		\begin{tabular}{|c|c|c|}
			\hline
			\# of Angles&\strucLearn&\baseLearn\\
			\hline
			2&$\mathbf{-6.8}$&7.3\\
			4&$\mathbf{-66}$&-14\\
			8&$\mathbf{-32}$&83.3\\	
			\hline
		\end{tabular}
		\end{center}
}
\caption{Test set negative log likelihood on the unsupervised Gait problem. Bold indicate performances that are significantly better than 
	non-bold.}
\label{tab:gaitDataUnSupervisedExpLikelihood}
\end{table}

In the conditional generation we experimented with one and two angles.
This time when it comes to one angle \strucLearn is significantly worse compared to \baseLearn. The situation is reversed
when we consider two angles. We hypothesize that the low performance in the one-angle setting is because most of the network 
representation power is consumed in learning and expressing correlations between the input features. With two angles we 
are able to learn and express correlations between the angles themselves, thus the better performance. 

As with the artificial dataset we also check the behavior of \strucLearn on the two angle dataset with respect to overfitting 
by visualising the evolution 
of the negative log likelihood in the training and testing set as a function of the number of epoch, right graph in figure~\ref{fig:simulatedDataSupervisedExpLikelihoodEvolutionAndgaitOverfitting}. 
As it was also the case with the artificial dataset we never observe a 
divergence between the performance on the training and the testing set, in fact here we even \strucLearn 
train for 1000 epoch, point to which we stopped without observing any 
divergence between the two losses. When it come to \baseLearn the overfitting is very severe 
and happens again around 100 epoch.

 	\begin{table}
	{
		\begin{center}
		\begin{tabular}{|c|c|c|c|c|c|c|c|}
			\hline
			Angles&\strucLearn $\lambda=0$ &\multicolumn{2}{|c|}{\strucLearn $\lambda=100$}&\multicolumn{2}{|c|}{\strucLearn $\lambda=10000$}&\multicolumn{2}{|c|}{RNN}\\
			\hline
			&mean&mean&p-value&mean&p-value&mean&p-value\\
			\hline
			1&-47&-44&$\mathbf{2\cdot 10^{-5}}$&-49&1.0&$\mathbf{-53}$&1\\
			2&$\mathbf{-125}$&-114&$\mathbf{0}$&-100&$\mathbf{0}$&-97&$\mathbf{0}$\\	
			\hline
		\end{tabular}
		\end{center}
	}
	\caption{Test negative log likelihood on the supervised	 Gait problem. Bold indicate the lowest statistically significative better negative log likelihood using  a $t$-test. }
	\label{tab:GaitSupervisedLikelihood}
\end{table}

\begin{figure}
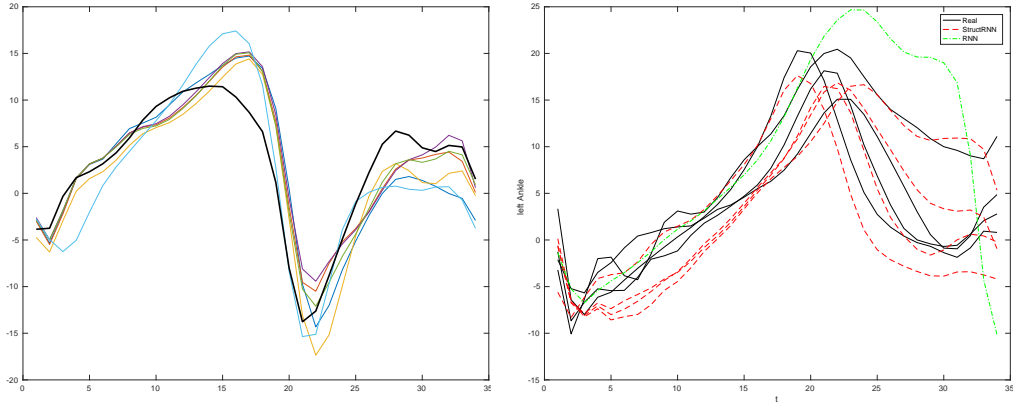

\begin{minipage}{0.49\columnwidth}
                \includegraphics[width=1\linewidth,trim={2cm 0 3.1cm 0},clip]{mult_state.pdf}
\end{minipage}
\begin{minipage}{0.49\columnwidth}
        \includegraphics[width=1\linewidth,trim={2cm 0 3.1cm 0},clip]{Supervised_Gait_Sample.pdf}
\end{minipage}
	\caption{Left graph: Sensitivity of output prediction to the serialisation order of $\mathbf x$. The thick black curve 
	is the real gait, the other curves are the ones generated from different perturbations of the $\mathbf x$ input. Right
	graph: Multiple gait cycles of a given patient (black), \strucLearn predictions generated from different serialisations of 
	the $\mathbf x$ of the given patient (red) and the \baseLearn prediction (green). }
        \label{fig:GaitCycleVarianceInputPermuationAndGaitCycleVarianceInputPermuationAndBaseline}
\end{figure}

To visualise the quality of predictions and how they are affected by the serialisation of $\mathbf x$ which we need to feed
to \strucLearn in order to generate the $\mathbf Y$ component we give in the left part of 
figure~\ref{fig:GaitCycleVarianceInputPermuationAndGaitCycleVarianceInputPermuationAndBaseline} 
the different predictions we get for one angle and the different permutations of the $\mathbf x$ vector. As we can see 
the predicted gait curves are globally consistent and rather similar to the true gait curve. Finally in 
right part of figure~\ref{fig:GaitCycleVarianceInputPermuationAndGaitCycleVarianceInputPermuationAndBaseline}
we give the multiple gait cycles of a single patient, the different predictions produced by \strucLearn using 
different serialisations of $\mathbf x$ and the prediction produced by \baseLearn. Again it is clear that the 
predictions generated by \strucLearn are much more consistent to the true gait structure compared to the ones 
generated by \baseLearn, which is considerably off from the true data structure.

\section{Hardware infrastructure}

The results in this paper were computed on a variety of hardware. As GPUs we used:
\begin{itemize}
	\item Geforce 980.
	\item Geforce 1070.
	\item Geforce 1080.
	\item Titan xp.
	\item P100.
	\item RTX 2080 ti.
\end{itemize}

As CPUs we used:
\begin{itemize}
	\item i7-5820K.
	\item i7-7700HQ.
	\item i9-9900K.
	\item i9-9820X.
	\item E5-2630.
\end{itemize}

The runtime of the experiments we present in the paper are 
at maximum of 2 days with a single gpu. In some of the preliminaries 
studies we did use longer running time.

\section{Model Complexity}
There are 4 components in our model that have a time complexity.
\begin{enumerate}
	\item The computation and back-propagation of the RNN.
	\item The computation and back-propagation of the Constraint regularizer on the state.
	\item The serialization of the instance together with the computation of the state.
	\item The computation of which states of the mini-batch are the same.
\end{enumerate}
The two back-propagation steps are done in training and cannot be pre-computed. They are in the critical path of the algorithm. Whereas the two last steps are pre-processing steps which can be computed asynchronously and in parallel. Theses steps are also computed on cpus. By using enough cores of cpus theses two last step have no influence in the training time.

The complexity of an RNN is linear with the length of the RNN which in our case is given by the complexity of a single data instance.
The complexity of the regularizer is also linear with the length of the serialization.

The complexity of the serialization of an instance was for all the structures we considered linear with the length of the serialization.

Finally, the complexity to find which states are equivalents is linear with the length of the serialization. However, to obtain this linear complexity we need to use an hash table on the set. Defining this hash function for every structure may not be easy.

In conclusion, the complexity of our model is not very different of the one of an RNN.

\section{Software infrastructure}

The first version of our code was implemented in torch\citep{torch} and c++\citep{c++17}.
The current version is based on PyTorch\citep{pytorch} and c++.
Additionally to manipulate the different datasets we used the following libraries:
\begin{itemize}
	\item PCL \citep{PCL} to manipulate point cloud data.
	\item RDKit \citep{rdkit} to read and export SMILES.
	\item deepchem\citep{Ramsundar-et-al-2019} to run the molecule baseline and export the SMILES.
	\item Sol2\citep{sol2} a lua wrapper to communicate between c++ and torch.
	\item Boost\cite{BoostLibrary} generic c++ tool to read and manipulate data.
\end{itemize}
Additionally to facilitate deployment on our clusters we used Docker, Kubernetes, Singularity and Shifter. 

\section{Code release}

We made the full code and datasets to run the Set and Graph experiment available at the following location \url{https://gitlab.com/nips6828Submission}.

The Set experiment is available at \url{https://gitlab.com/nips6828Submission/pointcloud} and the Graph experiment is available at \url{https://gitlab.com/nips6828Submission/molecule}.

For ease of use, we also published an official docker container (\url{www.docker.com/}) for both repositories.
To use the docker container you need to use a modern linux kernel ($>3.xx$) have an nvidia gpu with up to date driver ($>4xx$) and have nvidia-docker (\url{https://github.com/NVIDIA/nvidia-docker}).

Docker compatible variant like singularity, shifter or kubernetes can also be used.

Once the prerequisites are installed the image can be downloaded for the pointcloud dataset by:
\begin{verbatim}
docker pull registry.gitlab.com/nips6828submission/pointcloud:latest
\end{verbatim}
or for the molecule dataset by:
\begin{verbatim}
docker pull registry.gitlab.com/nips6828submission/molecule:latest
\end{verbatim}

Then you can enter the container with:
\begin{verbatim}
docker run --runtime=nvidia -it registry.gitlab.com/nips6828submission/pointcloud:latest
\end{verbatim}
or
\begin{verbatim}
docker run --runtime=nvidia -it registry.gitlab.com/nips6828submission/molecule:latest
\end{verbatim}

Once inside, the different binaries can be executed. Help about the options can be obtained by using the {\ttfamily--help} option.

Note that the release of theses containers is mainly for demonstration purpose. For real experiments it is recommended to store the dataset together with the result in a mounted folder.

In case of issues running the code an email can be send to \url{nips6828@gmail.com} or by posting an issue to the repository.

Upon acceptance of the paper the code will be published under our real name.

%
%
%
%
%

\end{document}